\documentclass[runningheads]{llncs}
\usepackage{graphicx}

\usepackage{tikz}
\usepackage{comment}
\usepackage{amsmath,amssymb} 
\usepackage{color}
\usepackage[english]{babel}
\usepackage{enumitem}
\usepackage{booktabs}
\usepackage{makecell}
\usepackage{blindtext}
\usepackage{xcolor-material}
\usepackage{subfig}

\usepackage[accsupp]{axessibility}  

\usepackage{amsmath}

\newcommand{\snamebig}[0] {POP\xspace}

\begin{document}

\setlength{\abovedisplayskip}{0pt}
\setlength{\belowdisplayskip}{0.2cm}
\setlength{\abovedisplayshortskip}{0pt}
\setlength{\belowdisplayshortskip}{0pt}

\let\cleardoublepage\clearpage
\pagestyle{headings}
\mainmatter
\def\ECCVSubNumber{5561}  

\title{POP: Mining POtential Performance of new fashion products via webly cross-modal query expansion}

\titlerunning{POP: Mining POtential Performance of new fashion products}
%

\newcommand\CoAuthorMark{\footnotemark[\arabic{footnote}]} 

\author{
Christian Joppi\thanks{indicates equal contribution.}\inst{1}\orcidID{0000-0003-4495-9515}
\and
Geri Skenderi\protect\CoAuthorMark\inst{2}\orcidID{0000-0001-9968-7727}
\and
Marco Cristani\inst{2,1}\orcidID{0000-0002-0523-6042}}

\authorrunning{C. Joppi et al.}
%
\institute{
Humatics s.r.l\\
\email{\{name.surname\}@sys-datgroup.com}\\
\and
University of Verona\\
\email{\{name.surname\}@univr.it}
}

\maketitle

\begin{abstract}
We propose a data-centric pipeline able to generate exogenous observation data for the New Fashion Product Performance Forecasting (NFPPF) problem, i.e., predicting the performance of a brand-new clothing probe with no available past observations. Our pipeline manufactures the missing past starting from a single, available image of the clothing probe. It starts by expanding textual tags associated with the image, querying related fashionable or unfashionable images uploaded on the web at a specific time in the past. A binary classifier is robustly trained on these web images by confident learning, to learn what was fashionable in the past and how much the probe image conforms to this notion of fashionability. This compliance produces the POtential Performance (POP) time series, indicating how performing the probe could have been if it were available earlier. POP proves to be highly predictive for the probe’s future performance, ameliorating the sales forecasts of all state-of-the-art models on the recent VISUELLE fast-fashion dataset. We also show that POP reflects the ground-truth popularity of new styles (ensembles of clothing items) on the Fashion Forward benchmark, demonstrating that our webly-learned signal is a truthful expression of popularity, accessible by everyone and generalizable to any time of analysis. Forecasting code, data and the POP time series are available at: \url{https://github.com/HumaticsLAB/POP-Mining-POtential-Performance}
\end{abstract}

\section{Introduction}
\label{sec:intro}
\begin{figure}[t!]
	\centering
	\small
	\includegraphics[width=\linewidth]{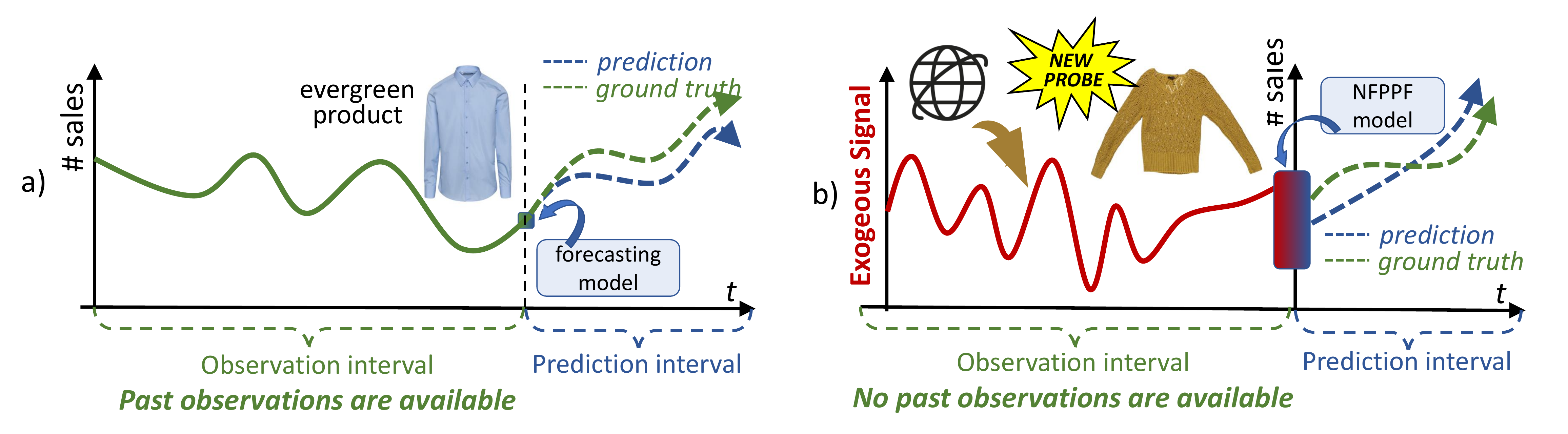} 
	\caption{\footnotesize a) A standard forecasting setup, where an evergreen item has past observations to exploit, \textit{e.g.,} \# sales; b) New Fashion Product Performance Forecasting (NFPPF) problem, where no past observations are available and \mbox{exogenous} data must be considered. Here we propose POP, the POtential Performance series, which is webly learned. The signals in b) appear on the same scale purely for visualization purposes, in reality this might not be the case.}
	\label{fig:opening_nfppf_deluxe}
\end{figure}

Forecasting the performance of a new clothing item is a crucial challenge for fashion companies~\cite{cheng2021fashion,fildes2019retail}. A good forecast in terms of predicted sales or product popularity can greatly help optimize the supply chain~\cite{ren2020demand} and minimize losses on multiple levels. Unfortunately, standard forecasting approaches require observations from the past to provide a forecast for the same product in the future~\cite{FPAP2,al2017fashion} and this information is typically available for evergreen products only (Fig.~\ref{fig:opening_nfppf_deluxe}a). In other cases, judgemental forecasts~\cite{FPAP2} from fashion professionals\footnote{A commercial example is Trendstop \url{https://www.trendstop.com/} and its ``Trend Platform Membership'' service} are the only ones that can help. Starting from photos or realistic renderings, which we call \emph{probe} images, they perform comparisons with trends as they surface and then infer the probe's success~\cite{ren2020demand}. 
In this paper, we try to model this line of reasoning and create a data-centric~\cite{motamedi2021data} pipeline that is able to extract a highly predictive signal from the web, dubbed ``POtential Performance'' (POP), which can be fed into any NFFPF forecasting model as an additional variable and lead to more accurate forecasts. (see  Fig.~\ref{fig:opening_nfppf_deluxe}b). 

Our cross-modal, query expansion based pipeline is sketched in Fig.~\ref{fig:Main.scheme2}. The input is a single probe image of the product to be analyzed, or a photorealistic rendering\footnote{Several such tools are available, for instance \url{https://www.tg3ds.com/3d-fashion-design-tools}.}. 
The pipeline first extracts textual tags from the probe automatically or by directly considering the associated technical sheet. The tag set is expanded with \emph{positive} and \emph{negative} tags that are used to perform a \emph{time-dependent} query online, i.e., collecting images of ``fashionable'' and ``unfashionable'' items related to the tags, which have been uploaded during some specified $K_{past}$ intervals in the past. These images are used to \emph{confidently learn}~\cite{northcutt2021confident} a binary classifier that captures what is fashionable VS unfashionable in that interval. This learning procedure prunes noisy images from both the positive and negative classes, resulting in a robust model. Subsequently, pruned positive images are projected into an embedding space by the learned model and compared with the (also projected) initial probe image, providing the $K_{past}$-long \snamebig signal. The \snamebig signal indicates how popular the probe could have been over time if it were available earlier in the past. 

Our approach should be cast in the field of \emph{data-centric artificial intelligence} (DCAI)~\cite{motamedi2021data}, since it automates the creation of high quality training data that can be used to improve any forecasting model which accommodates multivariate time series forecasting. \snamebig has been tested on diverse state-of-the-art NFPPF algorithms that predict sales curves of new products on the recent VISUELLE fast-fashion dataset~\cite{skenderi2021well}, providing superior performance when compared to other types of training signals. It has also been customized to deal with fashion styles (\textit{i.e.,} ensembles of clothing items) on the Fashion Forward benchmark~\cite{al2017fashion}. Fashion Forward (FF) calculates a popularity time series for an automatically extracted style based on the dataset properties and then applies standard forecasting algorithms. We substitute their popularity series with \snamebig, reaching similar predictions despite relying only on an exogenous input. Surprisingly, on the Dresses partition of FF, we reach the absolute best, suggesting that \snamebig can foresee the success of a potentially new fashion style. Summarizing, the contributions of this work are threefold:
\begin{enumerate}[noitemsep, leftmargin=*]
    \item The first data-centric strategy tailored to forecasting, used to create an exogenous observation signal which improves forecasts of the performance (number of sold items, popularity) of brand new clothing items with non-existent pasts. 
    \item A webly-learned method to freely collect information about fashion trends without relying on private or costly repositories.   
    \item Best overall results on all the tested NFPPF tasks.
\end{enumerate}

The rest of the paper is organized as follows: related literature is analyzed in Sec.~\ref{sec:related}; the proposed approach is detailed in Sec.~\ref{sec:method}; experiments are reported in Sec.~\ref{sec:experiments}, and finally; concluding remarks are drawn in Sec.~\ref{sec:conclusion}.

\section{Related literature}
\label{sec:related}
\textbf{NFPPF problem.} 
The NFPPF problem has been deeply investigated in the fields of quantitative fashion design~\cite{ren2017comparative,arvan2019integrating,jeon2020fashionq}, marketing and social sciences~\cite{silva2019googling,garcia2021fashion}, but is relatively new in the computer vision community. In both~\cite{ekambaram_attention_2020,singh_fashion_2019}, the main idea is that new products will sell comparably to similar, older products; this similarity is exploited in~\cite{singh_fashion_2019} via textual tags only, while in~\cite{ekambaram_attention_2020} an autoregressive RNN model takes past sales, textual product attributes, and the product image as input, to forecast the item sales. The work in~\cite{skenderi2021well} focuses on the additional direction of checking the past to look for predictive exogenous signals. In particular, the authors exploit Google Trends, querying textual attributes related to the probe and embed the resulting trend into a Transformer-based~\cite{vaswani2017attention} architecture, which considers images, text and other metadata. The authors also rendered accessible the first publicly available dataset for NFPPF, VISUELLE. In our paper we follow the idea of looking back to web data, but use images as the main representation of online fashionability, obtaining a richer exogenous signal. Predicting the success of new fashion styles has never been taken into account, with past works~\cite{al2017fashion,lo_dressing_2019,ma_knowledge_2020} focusing on the standard forecasting setup.

\textbf{Data-centric AI.} Data-Centric AI~\cite{motamedi2021data} (DCAI) shifts the attention from the models to the data used to train and evaluate them. It is a topic whose importance is constantly growing in many AI communities~\cite{anik2021data,northcutt2021pervasive,NG:2021}\footnote{\url{https://datacentricai.org/}.}, with important effects on CV \& ML. In general, DCAI investigates methodologies for accelerating open-source dataset creation from lower-quality resources. Consequently, it is tightly coupled with learning on noisy data, which aims at producing consistent and low noise data samples, or removing labeling noise and inconsistencies from existing data~\cite{northcutt2021confident,song2019selfie,wang2019symmetric}. Our methodology is data-centric, since it automates the creation of training data from a large amount of web resources, while removing labeling noise. Notably, it represents a novelty in the DCAI panorama, since it creates \emph{temporally-dependent} training data, i.e., time series, as it is required by NFPPF and in general by forecasting tasks.

\section{Methodology}
\label{sec:method}
\begin{figure}[t!]
	\centering
	\includegraphics[width=\linewidth]{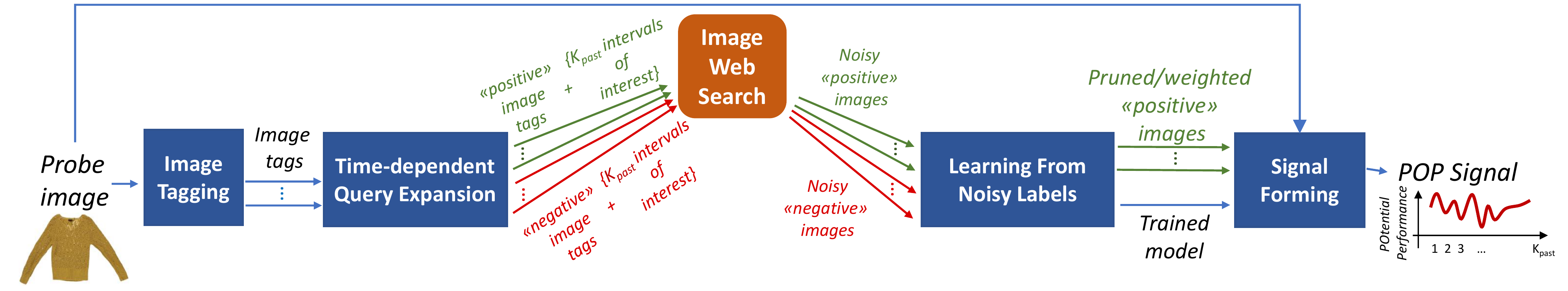}	
	\caption{\footnotesize Schematic pipeline of our approach; we start with a probe image and obtain the POtential Performance (\snamebig) signal at the end. Along this pipeline, we sequentially process information in different modalities, thereby creating a \textit{cross-modal signal}.}
	\label{fig:Main.scheme2}
\end{figure}

\sloppy The goal of our approach is to produce an exogenous variable that can aid a forecasting model in predicting the future performance of a product (sales, popularity). The input to our approach is the probe image $\mathbf{z}^{(t)}$, where $\mathbf{z}$ represents the new clothing item and $t$ the \emph{observation time}, which is the date from when we begin to look into the past. The output is the \snamebig signal $S^{(t)}_\mathbf{z}=s^{(t-K_{past})}_\mathbf{z},\ldots,s^{(t-k)}_{\mathbf{z}},\ldots,s^{(t-1)}_{\mathbf{z}}$, defined for $K_{past}$ time steps preceding $t$, where $k = 1, \ldots, K_{past}$ and $s^{(t-k)}_\mathbf{z} \in \mathbb{R}$. In this paper, we describe the observation times in terms of weeks and set $K_{past}=52$. This translates to looking one year prior to the observation time $t$, as typically done in fashion market analysis~\cite{sorger2017fundamentals}. The next sections will sequentially detail the general pipeline of our approach, depicted in Fig.~\ref{fig:Main.scheme2}.

\subsection{Image Tagging} The first operation is the extraction of textual tags $\{a_\mathbf{z}^{(j)}\}_{j=1,\ldots,J}$ associated to $\mathbf{z}$. These tags should represent the clothing item with sufficient generality, capturing at least categorical information (\textit{e.g} ``long sleeve'') and a dominant color (\textit{e.g}  ``yellow''). Empirically, we found these tags to work well while being easily obtainable. Category and color can be automatically extracted with high accuracy~\cite{liuLQWTcvpr16DeepFashion} or are usually contained in the technical data sheet accompanying the product, which is what we exploit in this work, as shown in Sec.~\ref{sec:experiments}.

\begin{figure}[t!]
	\centering
	\includegraphics[width=1\linewidth]{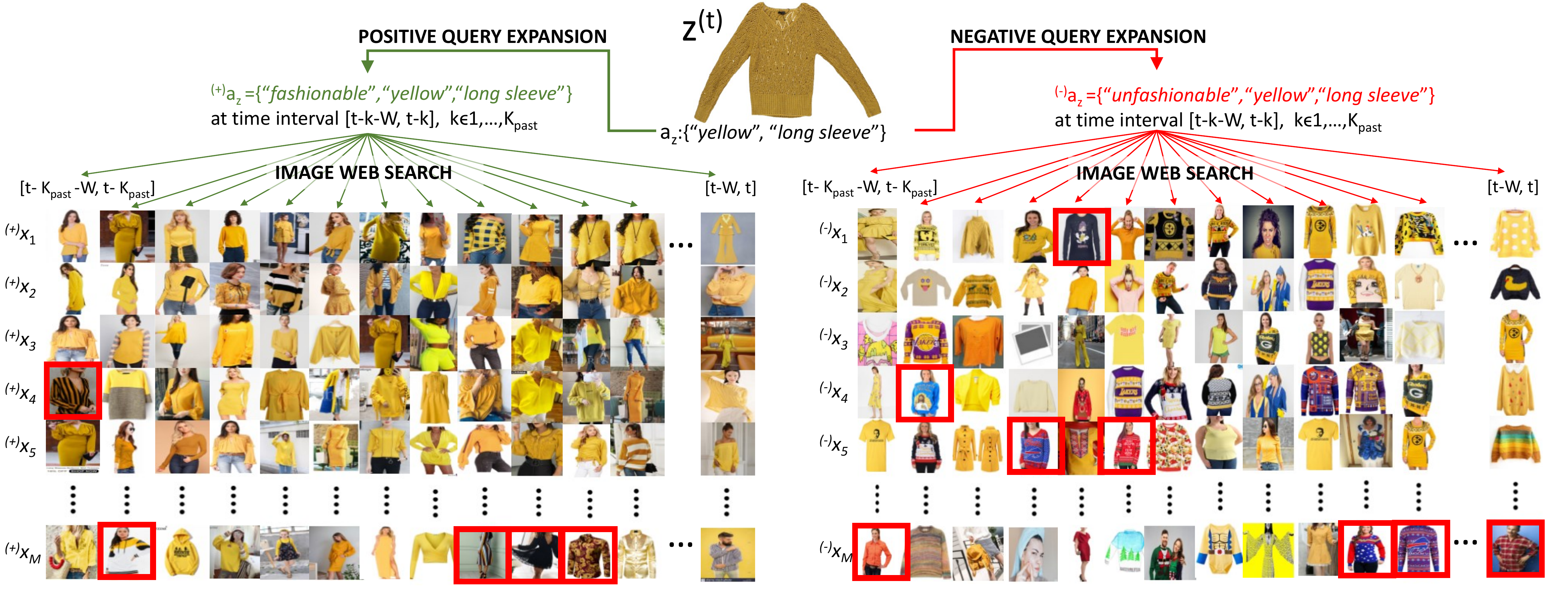}
	\caption{\footnotesize Pipeline insights on \emph{Time-dependent Query Expansion} (Sec.~\ref{sec:TDQE}), \emph{Image Web Search} (Sec.~\ref{sec:IWS}) and \emph{Learning From Noisy Labels} (Sec.~\ref{sec:LFNL}) steps. This figure reports a real world excerpt of the download and processing of $N$=2600 images ($N=2(M\times K_{past})$, $M$ = 25, $K_{past}$=52).}
	\label{fig:Image.analysis.ufficial}
\end{figure}

\subsection{Time-dependent Query Expansion}\label{sec:TDQE} The second operation (detailed in Fig.~\ref{fig:Image.analysis.ufficial} on a real example) performs two different textual query expansions, generating \emph{positive expansions}, ${\{a_\mathbf{z}^{(j)}\}}_{j=1,\ldots,J}\cup J^{(+)}$ where the additional $J^{(+)}$ tags indicate attractive clothing items, and conversely for \emph{negative expansions}. In this paper, we found the tags $J^{(+)}= $ ``fashionable'' and $J^{(-)}= $ ``unfashionable'' to be the most effective for positive and negative expansions, respectively. Alternatives as ``best seller'' and ``unattractive'' were considered, returning similar results. 

Each expansion, either positive or negative, is associated to a particular $k = 1, ...,  K_{past}$ for the time interval $[t-k-W,t-k]$, where $W$ is a temporal window we wish to consider for the image search, also expressed in weeks. In our experiments we set $W=4$, which translates to having a sliding window of size 4 and stride of 1 over the temporal axis. This allows the pool of downloaded images to disclose what are newly indexed items in relation to previous time steps, developing a temporal locality in the data pool. The precise value of $W$ was chosen after an empirical evaluation over the range $1, ..., 12$. 

\subsection{Image Web Search} \label{sec:IWS} A given expanded textual query along with a time interval is fed into a web API request to gather $M$ representative fashionable and unfashionable images ${}^{(+)}\{\mathbf{x}_i\}_{i=1,\ldots,M}^{(t-k)}; {}^{(-)}\{\mathbf{x}_i\}_{i=1,\ldots,M}^{(t-k)}$ that have been uploaded in the interval $[t-k-W,t-k]$, for $k=1,\ldots,K_{past}$. In particular, we adopt Google Image search,
selecting the first $M = 25$ images returned, assuming the ordering of Google Images perfectly mirrors a genuine image relevance~\cite{google_visualrank}. After the image web search phase, $M\times K_{past}$ fashionable and unfashionable images are collected respectively (as shown in Fig.~\ref{fig:Image.analysis.ufficial}). These images are then used to train a binary classifier $\theta$, aimed at distinguishing fashionable from unfashionable images. Webly learning and supervision based on Google Images has been considered before in computer vision, especially for image classification and object detection \cite{Fergus_googleimages_2005,chen_webly_2015,Li_webly_2021}. \snamebig goes one step further, merging visual and textual search while adding a time-dependent query expansion to create more discriminative image sets. Nevertheless, the labels assigned to the images from the query expansions might be noisy, therefore we apply a confident learning method.

\subsection{Learning From Noisy Labels}\label{sec:LFNL}
In the following, we adapt the confident learning (CL) methodology specifically for our binary problem. For a broader overview, readers may refer to~\cite{northcutt2021confident}. Let $\mathbf{X} = \{\mathbf{x}_i,\tilde{y_i}\}_{1 \ldots N}$ be our set of $N=2(M\times K_{past})$ images with associated observed noisy binary labels $\tilde{y_i} \in \{\text{``fashionable''},\text{``unfashionable''}\}$. CL assumes that a true, latent label $y^*_i \in \{\text{``fashionable''},\text{``unfashionable''}\}$ exists for every sample. CL requires two inputs: 1) the out-of-sample $N\times2$ matrix $\mathbf{\hat{P}}$ of predicted probabilities where $\mathbf{\hat{P}}_{i,h} = \hat{p}(\tilde{y_i}=h;\mathbf{x}_i,\mathbf{\theta})$ with $\mathbf{\theta}$ a generic (binary) classifier initially trained on $\mathbf{X}$; 2) the set of noisy labels $\{\tilde{y}_i\}$. Subsequently, a robust $2\times 2$ confusion matrix, called the \emph{confident joint} matrix $\mathbf{C}_{\tilde{y},y^*}$, is computed\footnote{We drop the index $i$ for clarity.}:

\begin{equation}
\begin{split}
 \mathbf{C}_{\tilde{y},y^*}(h,l)&=|\mathbf{\hat{X}}_{\tilde{y}=h,y^*=l}|, \text{with}  \\
 \mathbf{\hat{X}}_{\tilde{y}=h,y^*=l}&=\Bigg\{ \mathbf{x} \in \mathbf{X}_{\tilde{y}=h}:\hat{p}(\tilde{y}=l;\mathbf{x},\mathbf{\theta})\ge t_l \Bigg\}          \label{eq:cj}
\end{split}
\end{equation}
where $t_l$ is a threshold that represents the expected self confidence value for each class:
\begin{equation}
    t_l=\frac{1}{|\mathbf{X}_{\tilde{y}=l}|}\sum_{x\in \mathbf{X}_{\tilde{y}=l}}\hat{p}(\tilde{y}=l;x,\mathbf{\theta})
\end{equation}\label{eq:tj}
In practice, $ \mathbf{C}_{\tilde{y},y^*}$ counts only those elements which have been confidently classified in a particular class, where the term ``confident'' means with a probability that is higher than the average probability of an element belonging to that class. In simpler words, if samples labeled as belonging to class $h$ tend to have higher probabilities because the model is over-confident about class $h$, then $t_h$ will be proportionally larger.
It also worth noting that Eq.~\ref{eq:cj} corresponds to a simplified version of the general building procedure of the confident joint matrix $ \mathbf{C}_{\tilde{y},y^*}$ of ~\cite{northcutt2021confident}, which nonetheless in our case is perfectly acceptable since we deal with binary classification and no \emph{label collision} may happen, \textit{i.e.,} the fact that a noisy label can correspond to a more than a single alternative class. 

On this robust confusion matrix, we estimate label errors from the off diagonal elements of $\mathbf{C}_{\tilde{y},y^*}(h,l)$. 
Wrongly labeled images are therefore pruned (indicated by the red boxes in Fig.~\ref{fig:Image.analysis.ufficial}), obtaining the cleaned fashionable and unfashionable images ${}^{(+)}\{x'_i\}_{i=1,\ldots,M'^{(t-k)}}^{(t-k)}; {}^{(-)}\{x'_i\}_{i=1,\ldots,M''^{(t-k)}}^{(t-k)}$, where $M'^{(t-k)}$ and $M''^{(t-k)}$ indicate that we can have a different number of positive and negative images, respectively,  related to each $t-k$ time step, due to the noisy sample elimination. The classifier is retrained on the cleaned data, obtaining a robust trained model $\theta'$. This procedure is data-centric and model agnostic; the specific $\theta$ used in this work is described in Sec.~\ref{sec:experiments}.

\subsection{Signal Forming\label{sec:SF}}
The \snamebig signal $S^{(t)}_\mathbf{z}=s^{(t-K_{past})}_\mathbf{z},\ldots,s^{(t-k)}_{\mathbf{z}},\ldots,s^{(t-1)}_{\mathbf{z}}$, is computed by considering the cleaned fashionable images ${}^{(+)}\{\mathbf{x}'_{i}\}_{i=1,\ldots,M'^{(t-k)}}^{(t-k)}$, the robust model $\theta'$, and the image $\mathbf{z}$, as follows:
\begin{equation}
\small
s^{(t-k)}_{\mathbf{z}}=\frac{1}{M'^{(t-k)}}\sum_{i=1}^{M'^{(t-k)}} \frac{
\langle\theta'\left({}^{(+)}{\mathbf{x}'_{i}}^{(t-k)}\right) \cdot \theta'\left(\mathbf{z}\right)\rangle}
{
    {\parallel\theta'\left({}^{(+)}{\mathbf{x}'_{i}}^{(t-k)}\right)\parallel}
    {\parallel\theta'\left(\mathbf{z}\right)\parallel}
}
\label{eq:signal.forming}
\end{equation}

where $\theta'\left(\mathbf{z} \right)$ indicates the extracted features of $\mathbf{z}$ from $\theta'$, and  $\langle \cdot \rangle$ indicates the dot product. In other words, the signal value $s^{(t-k)}_{\mathbf{z}}$ is the average cosine similarity between the embedding of the probe image $\mathbf{z}$ and each fashionable image $\mathbf{x}'_{i^{(t-k)}}$ from the $M'^{(t-k)}$ downloaded images. An assessment of alternative signal forming options is shown in the supplementary material.

\section{Experiments}
\label{sec:experiments}
In line with the general requirements of DCAI~\cite{motamedi2021data}, we show how our automatically manufactured time series helps a forecasting model $\psi$ achieve better results on a given task $\gamma$. The main idea behind our approach is that by knowing \snamebig, the forecasting model can gain a context on the past which otherwise would be missing and therefore improve. To demonstrate this, we perform extensive evaluation on two tasks (and different forecasting models): \emph{new fashion product sales curve prediction}~\cite{singh_fashion_2019,ekambaram_attention_2020,skenderi2021well}, and \emph{style popularity forecasting}~\cite{al2017fashion}. We show ablative studies on the first task and an impressive outcome on the second.

The binary classifier $\theta$ for learning on noisy data (Sec.~\ref{sec:LFNL}) is based on a ResNet50~\cite{he2015deep}, pre-trained on ImageNet~\cite{imagenet}, with two additional fully connected layers. During the confident learning procedure, we fine-tune its last convolutional block and fully connected layers for 50 epochs with a batch size of 64, using CE loss, following a 5-fold cross validation protocol. AdamW~\cite{loshchilov2018decoupled} is used as optimizer, with a learning rate of $1e-4$. The forecasting neural network models are all trained for 200 epochs with a batch size of 128 and L2 loss, using the AdaFactor~\cite{shazeer2018adafactor} optimizer. The experiments are performed on two NVIDIA 3090 RTX GPUs. 

\subsection{Task 1: New Fashion Product Sales Curve Prediction}\label{subsec:task1}
The output of a sales curve forecasting model for a probe clothing item $\mathbf{z}$ is a time series
$O^{(st)}_\mathbf{z}=o^{(st+1)}_\mathbf{z},\ldots,o^{(st+k)}_\mathbf{z},\ldots,o^{(st+K_{fut})}_\mathbf{z}$ that indicates how many pieces of $z$ will be sold starting at a particular time step $st$ (typically the start of the season), for the next $K_{fut}$ time steps.

We run our first set of experiments on the VISUELLE dataset~\cite{skenderi2021well}. For each available product, multi-modal information is provided: i) images, ii) text tags, iii) Google Trends, iv) sales curves. The evaluation protocol follows that of VISUELLE, simulating how a fast-fashion company deals with new products on two particular moments: the \emph{first order setup} and the \emph{release setup}. The former takes place when the company decides which products and how many pieces to order by looking at probe images. The latter is right before the season, and is useful to obtain an accurate forecast in order to plan stock replenishment. These two setups use 28 and 52 week long exogenous signals (originally Google Trends~\cite{skenderi2021well}), respectively.

Note that for the sake of fairness, we do not alter the training setup or models from \cite{skenderi2021well}, keeping the cardinality and the type of the training data fixed and \emph{substituting only the Google Trends with our \snamebig signal}. All the models are trained considering the 12-week long sales signals, whilst the evaluation is done on a 6-week horizon. This is shown to give the best predictions while simulating politics of real fashion companies~\cite{skenderi2021well}.

\begin{figure}
    \centering
    \includegraphics[width=\linewidth]{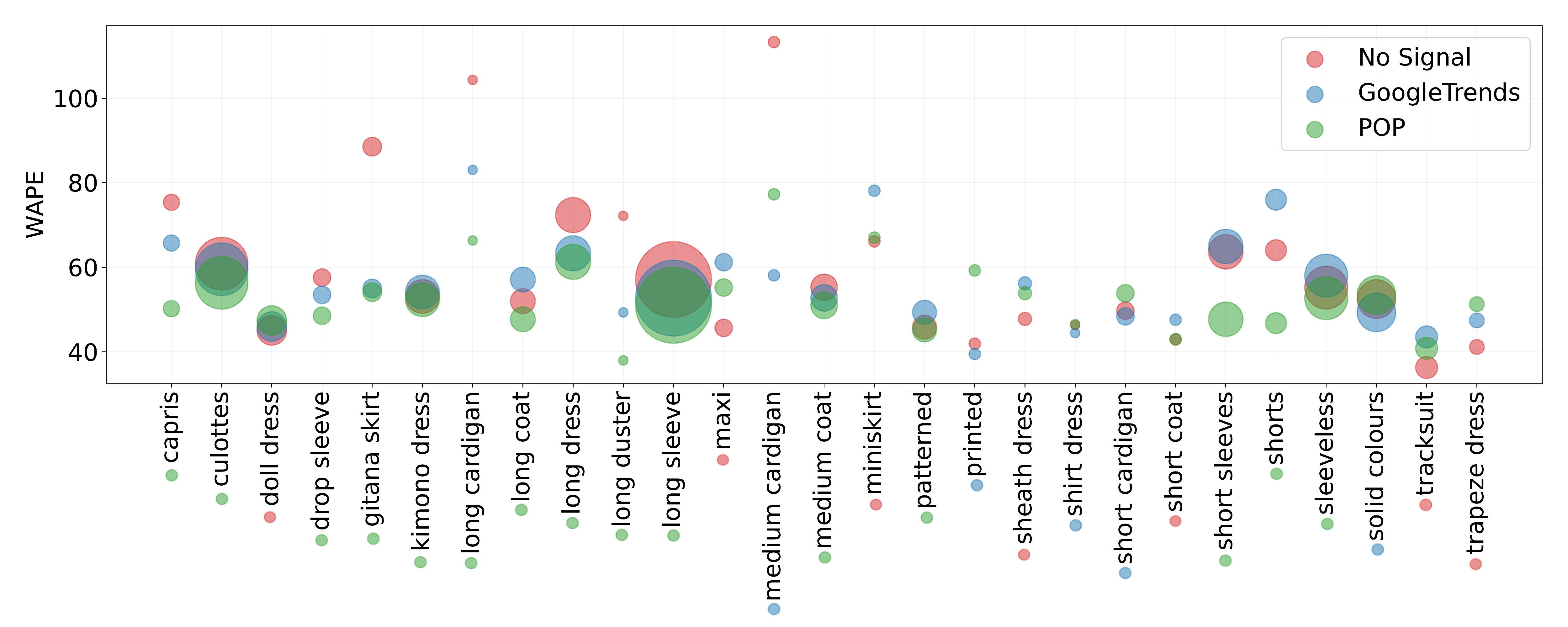}
    \caption{\footnotesize Forecasting WAPE results per clothing category; the larger the blob, the higher the \# of items in that category; the color below each category name indicates the type of training setup which gives the best WAPE.}
    \label{fig:result_categories}
\end{figure}

We consider 5 algorithms (from oldest to newest): \emph{Gradient Boosting} for forecasting~\cite{ilic2021explainable}, Concat Multi-Modal RNN~\cite{ekambaram_attention_2020}  (\emph{Concat MM RNN} in the tables),  Residual Multi-Modal RNN~\cite{ekambaram_attention_2020} (\emph{Residual MM RNN}), Cross-Attention RNN~\cite{ekambaram_attention_2020} (\emph{X-Attention RNN}) and GTM Transformer~\cite{skenderi2021well} (\emph{GTM Transf.}) We consider the Weighted Absolute Percentage Error (WAPE) as primary evaluation metric and also compute the Mean Absolute Error (MAE) to demonstrate the error on an absolute scale~\cite{skenderi2021well}:


Note that the WAPE is not bounded by 100. Finally, we measure the similarity of the slope of the predicted curve with the ground truth using the \emph{Edit distance with Real Penalty} (ERP)~\cite{chen2004marriage}. This metric counts the number of edit operations (insert, delete, replace) that are necessary to transform one series into the other. Because we are dealing with continuous values, a threshold $\epsilon$=0.03 is used to decide if values are considered different and have to be edited.

\begin{table}[t]
\centering
\caption{\footnotesize\label{table:visuelle_results_fo}Results on VISUELLE with the \emph{first order setup}; ``W'' stands for WAPE, ``M'' for MAE. Lower is better for all metrics.}
\resizebox{\linewidth}{!}{
\begin{tabular}{|l|c | c | c | c |c |c |c |c |c |c |c |c |c |c |c |}\hline 
     \multicolumn{16}{|l|}{\hfil\textbf{\emph{First Order Setup ($K_{best}$ = 28 weeks)}}} \\ \hline
    \textbf{Exogenous} 
    &  \multicolumn{3}{l|}{\hfil\textbf{\emph{Gradient}}}
    &  \multicolumn{3}{l|}{\hfil\textbf{\emph{Concat}}}
    &  \multicolumn{3}{l|}{\hfil\textbf{\emph{Residual}}}
    &  \multicolumn{3}{l|}{\hfil\textbf{\emph{X-Attention}}}
    &  \multicolumn{3}{l|}{\hfil\textbf{\emph{GTM}}}\\
    
    \textbf{Signal}
    &  \multicolumn{3}{l|}{\hfil\textbf{\emph{Boosting}}}
    &  \multicolumn{3}{l|}{\hfil\textbf{\emph{MM RNN}}}
    &  \multicolumn{3}{l|}{\hfil\textbf{\emph{MM RNN}}}
    &  \multicolumn{3}{l|}{\hfil\textbf{\emph{RNN}}}
    &  \multicolumn{3}{l|}{\hfil\textbf{\emph{Transformer}}}\\

    &  \multicolumn{3}{l|}{\hfil\textbf{\cite{ilic2021explainable} 2020}}
    &  \multicolumn{3}{l|}{\hfil\textbf{\cite{ekambaram_attention_2020} 2020}}
    &  \multicolumn{3}{l|}{\hfil\textbf{\cite{ekambaram_attention_2020} 2020}}
    &  \multicolumn{3}{l|}{\hfil\textbf{\cite{ekambaram_attention_2020} 2020}}
    &  \multicolumn{3}{l|}{\hfil\textbf{\cite{skenderi2021well} 2021}}\\ 
    &W & M & ERP&W & M & ERP&W & M& ERP&W & M& ERP&W & M& ERP\\\hline
    
\emph{\makecell{No\\Signal}}&64.10&35.02&0.43&63.31&34.41&0.42&64.26&34.92&0.44&59.49&32.33&0.38&56.62&30.93&0.37\\ \hline
\makecell{Google\\Trends}&64.29&35.12&0.43&64.11&34.84&0.43&68.11&37.02&0.47&58.70&31.90&0.38&56.83&31.05&0.35\\ \hline
\textbf{\makecell{POP\\Signal}}&\textbf{63.75}&\textbf{34.83}&\textbf{0.42}&\textbf{58.09}&\textbf{31.73}&\textbf{0.39}&\textbf{58.88}&\textbf{32.16}&\textbf{0.39}&\textbf{57.78}&\textbf{31.56}&\textbf{0.38}&\textbf{53.41}&\textbf{29.18}&\textbf{0.32}\\ \hline
\end{tabular}}

\end{table}

\begin{table}[t]
\centering
\caption{\footnotesize \label{table:visuelle_results_re}Results on VISUELLE with the \emph{release setup}; ``W'' stands for WAPE, ``M'' for MAE. Lower is better for all metrics.}
\resizebox{\linewidth}{!}{
\begin{tabular}{|l|c | c | c | c |c |c |c |c |c |c |c |c |c |c |c |}\hline 
     \multicolumn{16}{|l|}{\hfil\textbf{\emph{Release Setup ($K_{best}$ = 52 weeks)}}} \\ \hline
    \textbf{Exogenous} 
    &  \multicolumn{3}{l|}{\hfil\textbf{\emph{Gradient}}}
    &  \multicolumn{3}{l|}{\hfil\textbf{\emph{Concat}}}
    &  \multicolumn{3}{l|}{\hfil\textbf{\emph{Residual}}}
    &  \multicolumn{3}{l|}{\hfil\textbf{\emph{X-Attention}}}
    &  \multicolumn{3}{l|}{\hfil\textbf{\emph{GTM}}}\\
    
    \textbf{Signal}
    &  \multicolumn{3}{l|}{\hfil\textbf{\emph{Boosting}}}
    &  \multicolumn{3}{l|}{\hfil\textbf{\emph{MM RNN}}}
    &  \multicolumn{3}{l|}{\hfil\textbf{\emph{MM RNN}}}
    &  \multicolumn{3}{l|}{\hfil\textbf{\emph{RNN}}}
    &  \multicolumn{3}{l|}{\hfil\textbf{\emph{Transformer}}}\\

    &  \multicolumn{3}{l|}{\hfil\textbf{\cite{ilic2021explainable} 2020}}
    &  \multicolumn{3}{l|}{\hfil\textbf{\cite{ekambaram_attention_2020} 2020}}
    &  \multicolumn{3}{l|}{\hfil\textbf{\cite{ekambaram_attention_2020} 2020}}
    &  \multicolumn{3}{l|}{\hfil\textbf{\cite{ekambaram_attention_2020} 2020}}
    &  \multicolumn{3}{l|}{\hfil\textbf{\cite{skenderi2021well} 2021}}\\ 

    &W & M & ERP&W & M & ERP&W & M& ERP&W & M& ERP&W & M& ERP\\\hline
    
\emph{\makecell{No\\Signal}}&64.10&35.02&0.43&63.31&34.41&0.42&64.26&34.92&0.44&59.49&32.33&0.38&56.62&30.93&0.37\\ \hline
\makecell{Google\\Trends}&63.52&34.70&0.42&65.87&35.80&0.44&68.46&37.21&0.48&59.02&32.08&0.38&55.24&30.18&0.33\\ \hline
\textbf{\makecell{POP\\Signal}}&\textbf{63.38}&\textbf{34.62}&\textbf{0.42}&\textbf{57.43}&\textbf{31.37}&\textbf{0.36}&\textbf{58.38}&\textbf{31.89}&\textbf{0.39}&\textbf{57.36}&\textbf{31.33}&\textbf{0.36}&\textbf{52.39}&\textbf{28.62}&\textbf{0.29}\\ \hline
\end{tabular}}

\end{table}

The results are shown in Table~\ref{table:visuelle_results_fo} for the \emph{first order setup} and in Table~\ref{table:visuelle_results_re} for the \emph{release setup}. 
As reference, we also report results \emph{without} any exogenous series, to show the net value of these indicators. For all the algorithms and both setups, adding \snamebig to the model boosts the performances over all the metrics, reaching the absolute best when coupled with GTM Transformer. On average, in the \emph{first order setup}, we improve the WAPE by 3.42\% over the Google Trends and by 3.21\% over not using any exogenous signals. In the \emph{release setup}  we improve by 2.85\% over the Google Trends and by 4.23\% over not using exogenous signals. These results demonstrate how our data-centric approach can provide optimal forecasts by creating a highly-predictive signal of past popularity that is image-based, unlike Google Trends. The forecasts are performed on 497 products over different stores, meaning that these improvements can provide a large impact on the supply chain operations.

\begin{figure}
    \centering
    \includegraphics[width=0.49\linewidth]{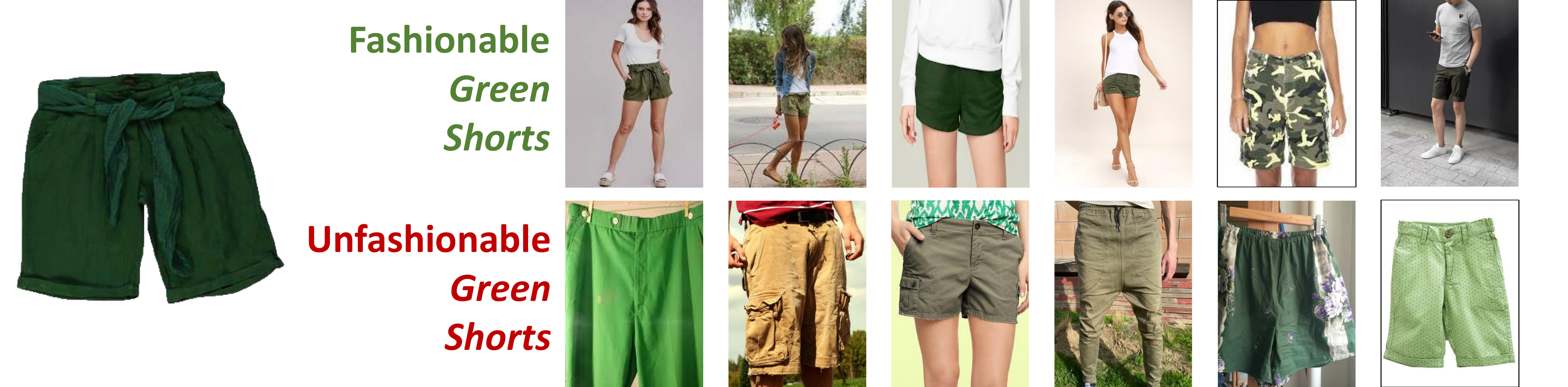}
    \includegraphics[width=0.49\linewidth]{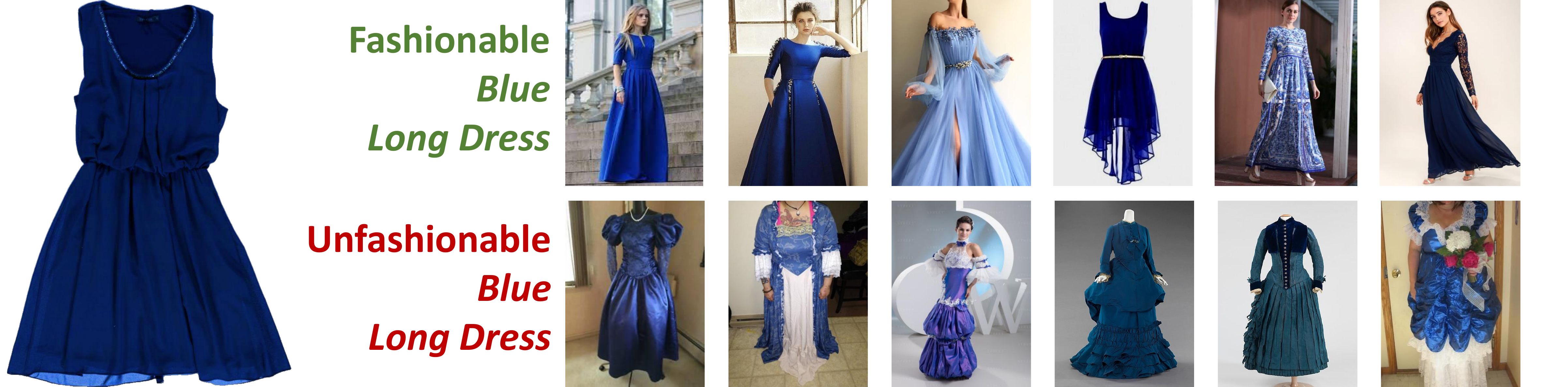}
    \includegraphics[width=0.49\linewidth]{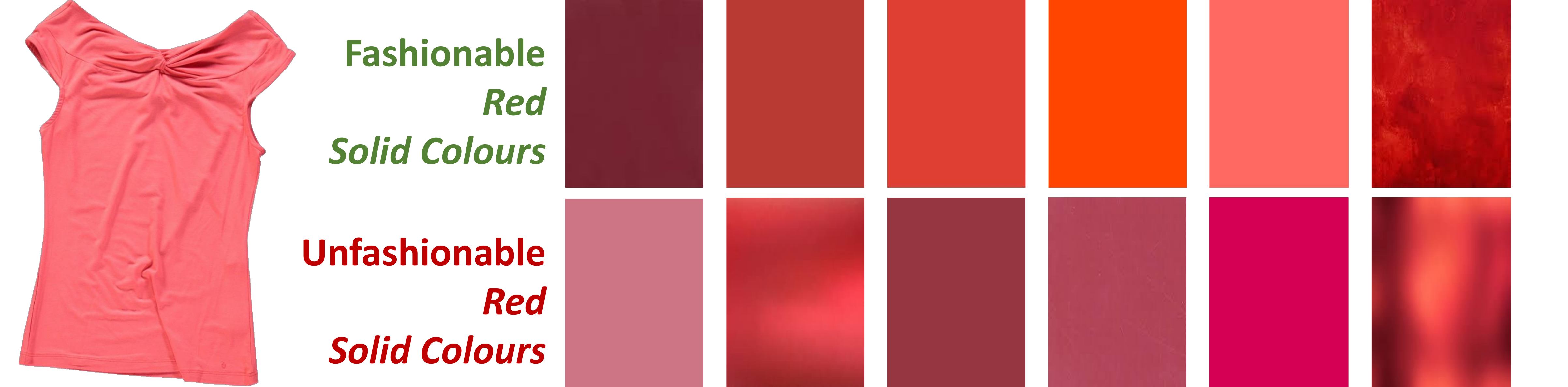}
    \includegraphics[width=0.49\linewidth]{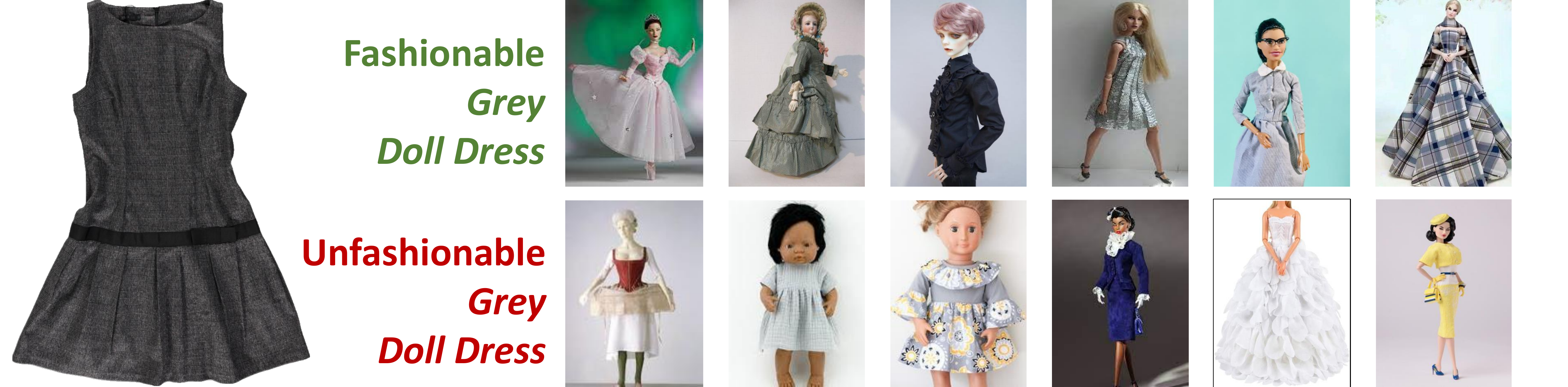}
    \caption{\footnotesize \label{fig:pop_images_examples}Examples of VISUELLE items (seasons SS17 and SS18 on the left, SS19 and AI19 on the right, respectively) and the correspondent fashionable/unfashionable images from the web. 
    Some web images can be misleading, due to the questionable category names of the VISUELLE dataset (``solid colours'', ``doll dress'').}
    \vspace{-1cm}
\end{figure}

In Fig.~\ref{fig:result_categories} we show the WAPE \emph{per clothing category}. We mostly perform better than the other training alternatives, yet some particular categories display limitations of our approach. These limitations arise due to the fact that the Image Tagging phase is assumed as flawless, since we rely on the technical sheet accompanying the probe image to extract the tags. The results per category (Fig.~\ref{fig:result_categories}) display how possibly mislabeled categories, or categories labeled in a general manner (``solid colours'',``doll dress'') may lead to misleading web images. As visible in Fig.~\ref{fig:pop_images_examples}, the related images from the web, both fashionable and not, are completely useless, since the tag of the category itself is misleading. In such cases, a robust automated category extraction could potentially lead to better results.

\subsubsection{Ablation studies.}\label{sec:alternative}
In the following, we focus on alternative versions of our proposed pipeline, ablating the specific modules illustrated in Fig.~\ref{fig:Main.scheme2}.
Table~\ref{tab:alternative_pipelines} contains all the results.\\
\noindent\textbf{Time dependent query expansion}
\begin{itemize} [noitemsep, leftmargin=*]
    \item{\emph{No expansion:}} Images are queried with the original tags collected in the Image Tagging phase, without generating positive or negative expansions. This is equivalent to querying only with "color + category". The learning step is impacted directly, since no positive or negative classes are available for learning, therefore we use our backbone model to extract image features. For each image $\mathbf{z^{(t)}}$, the web images $\{\mathbf{x}_i\}_{i=1,\ldots, M}^{(t-k)}$ that have been uploaded in the interval $[t-k-W,t-k]$, for $k=1,\ldots,K_{past}$ are collected. The signal forming Eq.~\ref{eq:signal.forming} changes accordingly, using all the $M$ downloaded images;
    \item{\emph{Misaligned past:}} We modify the query expansions by looking one year earlier than the ``correct'' past. Given the observation time $t$ of the probe $\mathbf{z}^{(t)}$, instead of looking backwards from $t-1$ weeks to $t-K_{past}$, we go from $t-1-K_{past}$ to $t-2\cdot K_{past}$.
    \vspace{-0.1cm}
\end{itemize}
With respect to all the alternative versions in this study, the \emph{No expansion} ablation gives the worst result. \snamebig provides an improvement of 0.73\% and 1.06\% WAPE for the \emph{first order setup} and \emph{release setup}, respectively.   
The \emph{Misaligned past} yields slightly better results, but still performs worse than \snamebig by 0.63\% and 0.22\% WAPE for the \emph{first order setup} and \emph{release setup}, respectively. This confirms that fashion has an evolution that changes year after year that we have to take into account.

\begin{table}[t]
\scriptsize
    \centering
    \caption{\footnotesize \label{tab:alternative_pipelines}Alternative versions of our pipeline (Fig.~\ref{fig:Main.scheme2}) on both the \emph{release} and \emph{first order } setups; ``W'' stands for WAPE, ``M'' for MAE. Lower is better for all metrics.}
    \setlength\extrarowheight{1pt}
    \resizebox{0.7\linewidth}{!}{
    \begin{tabular}{|l||c|c||c|c|}
        \hline
        \multicolumn{5}{|l|}{\hfil\textbf{\emph{Time Dependent Query Expansion}}}\\ \hline
        &\multicolumn{2}{l||}{\hfil\textbf{\emph{Release Setup}}}&\multicolumn{2}{l|}{\hfil\textbf{\emph{First Order Setup}}} \\
        \textbf{Strategy} & \textbf{W} & \textbf{M} & \textbf{W} & \textbf{M} \\ \hline
        \emph{No Expansion}&53.12&29.02&54.47&29.77\\ \hline
        \emph{Misaligned past}&53.02&28.96&53.63&29.30\\ \hline
        \multicolumn{5}{|l|}{\hfil\textbf{\emph{Learning With Noisy Labels}}} \\ \hline
         &\multicolumn{2}{l||}{\hfil\textbf{\emph{Release Setup}}}&\multicolumn{2}{l|}{\hfil\textbf{\emph{First Order Setup}}} \\
       \textbf{Strategy} & \textbf{W} & \textbf{M}& \textbf{W} & \textbf{M} \\ \hline
        \emph{No Learning}&53.03&28.97&53.83&29.41\\ \hline
        \emph{No Robust Learning}&52.81&28.85&53.59&29.28\\ \hline
        \emph{Symmetric Cross Entropy}~\cite{wang2019symmetric}&52.63&28.75&53.58&29.27\\ \hline
        \emph{SELFIE}~\cite{song2019selfie}&52.56&28.71&53.51&29.23\\ \hline
        \textbf{POP}&\textbf{52.39}&\textbf{28.62}&\textbf{53.41}&\textbf{29.18}\\ \hline
    \end{tabular}}
\end{table}

\vspace{0.1cm}
\noindent\textbf{Learning from noisy data}
\begin{itemize} [noitemsep, leftmargin=*]
    \item{\emph{No learning}}: A predefined image classification network is used to compute the distance among embeddings of the probe image with the positive, downloaded images. This is equivalent to ablating the ``Learning from Noisy Data'' phase of Fig.~\ref{fig:Main.scheme2}. It will highlight the importance of dealing with distances among embeddings which are specifically learned against distances coming from a general purpose network. We utilise the backbone of our binary classifier, specified in the introduction of Sec.~\ref{sec:experiments};
    \item{\emph{No robust learning}}: All of the downloaded positive and negative images are used to learn our binary classifier without pruning noisy data by confident learning;
    \item{\emph{Symmetric cross entropy}~\cite{wang2019symmetric}}: SCE is a robust classification loss; it adds to the standard cross entropy loss a \emph{reverse cross entropy} term which assumes the predicted labels as ground truth, and the original labels as possibly faulty. In practice, it penalizes noisy labels, without removing any associated training data;

    \item{\emph{SELFIE}~\cite{song2019selfie}}: the key idea is to correct the label of noisy \emph{refurnishable} samples with high precision, with the help of clean data which is defined as those samples within a mini-batch creating a small loss. Repeated training runs (dubbed ``restarts'') allow to use more training data, \textit{i.e.,} noisy samples which have been corrected in their labels. In particular, we use 3 restarts, after which 1.1\% of both fashionable and unfashionable items have been removed from the training data.
 \end{itemize}
The results in Table~\ref{tab:alternative_pipelines} show slightly different performances, promoting the general idea of learning from webly data. \emph{No learning} gives the worse performance, indicating that a fine tuning on the web data is beneficial (53.03 and 53.83 WAPE); when learning is done on the web data, there is some increase (52.81 and 53.59 WAPE); when learning is robust to label noise, with SCE, performances are better (52.63 and 53.58 WAPE); removing some outliers with SELFIE gives a further help (52.56 and 53.51 WAPE). Confident learning  remains the best solution, with 52.39 and 53.41 WAPE, while removing 0.8\% and 1.1\% of fashionable and unfashionable items respectively, from the 14,500,200 images mined using our cross-modal pipeline.

\begin{figure}
    \centering
    \includegraphics[width=0.7\linewidth]{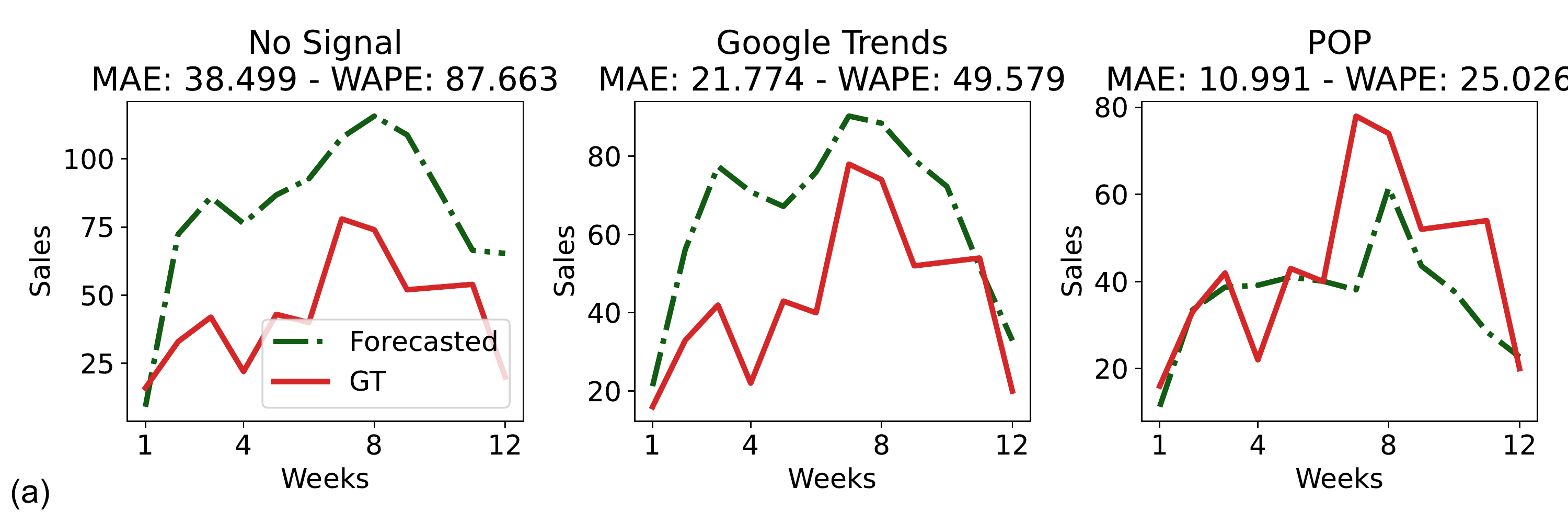}
    \includegraphics[width=0.7\linewidth]{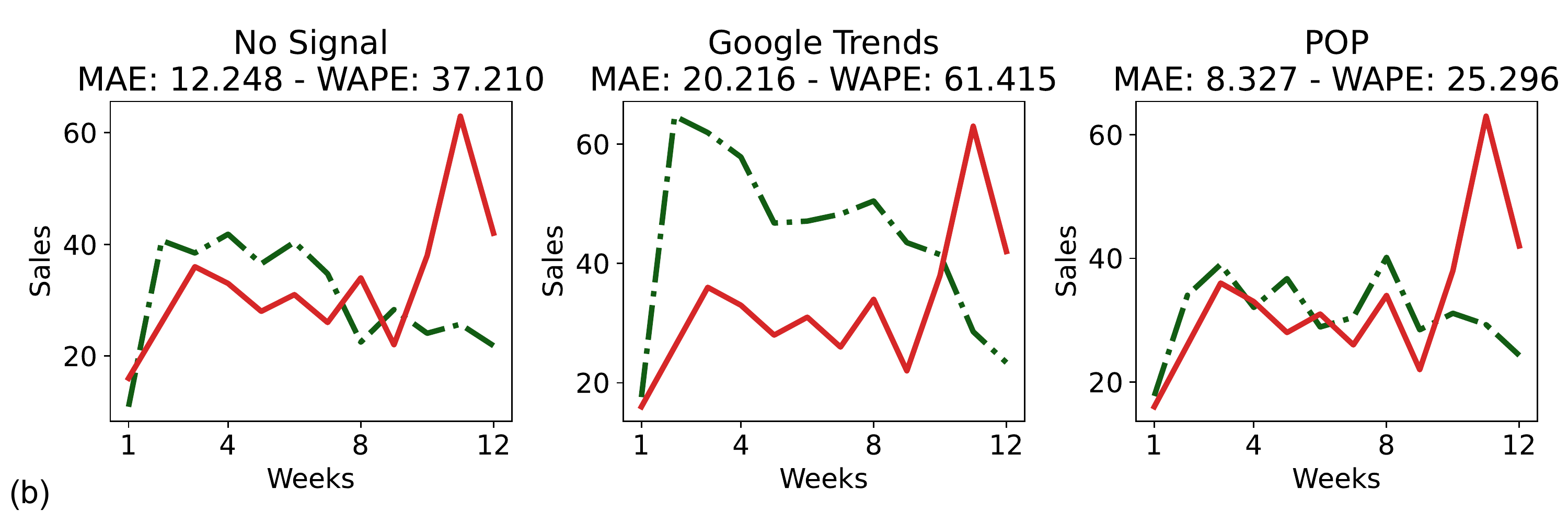}
    \caption{\footnotesize \label{fig:qlt_visuelle} Qualitative results for the sales forecast of two different products on VISUELLE, considering all 12 time-steps. In all cases, using POP provides better forecasts. In the bottom row (bottom right plot), we show a forecasting failure case, where the product is discounted in its final week of sales.}
    \vspace{-1cm}
\end{figure}

\subsection{Task 2: Popularity Prediction Of Fashion Styles}\label{sec:task2FF}
The style popularity prediction task~\cite{al2017fashion} is different from product sales forecasting in that it considers a popularity signal $y$ based on multiple clothing items. In the literature~\cite{al2017fashion,ma2020kern}, style is defined as a latent property of a set of clothing images that share some common visual features. Concretely, in Fashion Forward (FF)~\cite{al2017fashion}, Non-negative Matrix factorization is applied to extract $K$ styles from the attribute extraction features~\cite{liuLQWTcvpr16DeepFashion} of all the product images. Formally, let $\mathbf{A} \in \mathbb{R}^{M \times N}$ indicate the confidence that each of the M visual attributes is contained in each of the N images. $\mathbf{A}$ can be factorized into two matrices with non-negative entries:
\begin{equation}
    \mathbf{A} \approx \mathbf{W}\mathbf{H}, \mathbf{W} \in \mathbb{R}^{M\times K}\quad\textrm{and}\quad\mathbf{H} \in \mathbb{R}^{K\times N} 
\end{equation}
where $\mathbf{W}$ represents the confidence that each attribute is part of a style and $\mathbf{H}$ represents the confidence that each style is associated to an image. The popularity signal $y$ for a style $k$ is built by considering the interactions in the Amazon Reviews dataset~\cite{2015_amazonreviews} of all the items $\{\mathbf{z}\} \in A$ at time $t$, weighted by their style membership $H(k, z)$. For a detailed explanation, we refer to~\cite{al2017fashion}.

\begin{table}
    \centering
    \caption{\footnotesize Average results over all Fashion Forward~\cite{al2017fashion} dataset partitions and specific results for the Dresses partition, where \snamebig outperforms even the original GT style popularity time series (\emph{Oracle}).}
    \label{tab:ff_res}
    \resizebox{\linewidth}{!}{
    \begin{tabular}{|l|c c|c c|c c|c c|c c|c c|}
    \hline
    \multicolumn{13}{|l|}{\hfil\textbf{Global Average}}\\
    \hline
    \textbf{Signals}&\multicolumn{2}{l|}{\hfil\textbf{Mean}}&\multicolumn{2}{l|}{\hfil\textbf{Last}}&\multicolumn{2}{l|}{\hfil\textbf{Drift}}&\multicolumn{2}{l|}{\hfil\textbf{AR}}&\multicolumn{2}{l|}{\hfil\textbf{ARIMA}}&\multicolumn{2}{l|}{\hfil\textbf{SES}}\\
    &MAE&MAPE&MAE&MAPE&MAE&MAPE&MAE&MAPE&MAE&MAPE&MAE&MAPE\\ \hline
    \emph{Oracle}&\emph{0.136}&\emph{0.170}&\emph{0.093}&\emph{0.114}&\emph{0.174}&\emph{0.222}&\emph{0.271}&\emph{0.403}&\emph{0.136}&\emph{0.167}&\emph{0.094}&\emph{0.116}\\\hline
    GoogleTrends&0.846&1.000&0.846&1.000&0.846&1.000&0.846&1.000&0.846&1.000&0.846&1.000\\ \hline
    POP&0.152&0.192&0.116&0.144&0.182&0.229&0.281&0.418&0.235&0.293&0.125&0.156\\\hline\hline
    
    \multicolumn{13}{|l|}{\hfil\textbf{Dresses}}\\
    \hline
    \textbf{Signals}&\multicolumn{2}{l|}{\hfil\textbf{Mean}}&\multicolumn{2}{l|}{\hfil\textbf{Last}}&\multicolumn{2}{l|}{\hfil\textbf{Drift}}&\multicolumn{2}{l|}{\hfil\textbf{AR}}&\multicolumn{2}{l|}{\hfil\textbf{ARIMA}}&\multicolumn{2}{l|}{\hfil\textbf{SES}}\\
    &MAE&MAPE&MAE&MAPE&MAE&MAPE&MAE&MAPE&MAE&MAPE&MAE&MAPE\\ \hline
    \emph{Oracle}&\emph{0.155}&\emph{0.197}&\emph{0.130}&\emph{0.158}&\emph{0.203}&\emph{0.263}&\emph{0.307}&\emph{0.409}&\emph{0.173}&\emph{0.209}&\emph{0.129}&\emph{0.157}\\\hline
    GoogleTrends&0.849&1.000&0.849&1.000&0.849&1.000&0.849&1.000&0.849&1.000&0.849&1.000\\\hline
    \textbf{POP}&\textbf{0.119}&\textbf{0.157}&\textbf{0.108}&\textbf{0.127}&\textbf{0.173}&\textbf{0.216}&\textbf{0.229}&\textbf{0.334}&\textbf{0.162}&\textbf{0.193}&\textbf{0.109}&\textbf{0.130}\\\hline
    \end{tabular}}  

\end{table}

To extend this problem to a NFPPF setup, we have to imagine we are evaluating the performance of a brand new style that does not have a past. The purpose of \snamebig becomes replacing the original style popularity series. This means that POP has to be modified to deal with a style and not with a single clothing item, where two challenges are presented: 1) To verify how similar \snamebig is to the ground truth popularity signal and; 2) To check if \snamebig is highly predictive of the future popularity.


To deal with the first challenge, we consider for each style $k$ the 2 textual attributes~\cite{liuLQWTcvpr16DeepFashion} $w_1, w_2$ (extracted from $\mathbf{W}$) with the highest confidence scores and use them for the time dependent query expansion. FF provides the only dataset for style forecasting where both images and product metadata are available. The task is to predict a popularity score on a yearly basis. The data ranges from $[2008-2013]$, but since Google Images returns little to no images for queries before 2010, we use the range $[2010-2013]$ in our experiments. We set $K_{past} = 208$, meaning that we investigate 4 years back. In this way we can create a weekly series for each year and use the average as the value representing the popularity for that year. As probe image to create our \snamebig signal, we consider the top 10 images $\{\mathbf{z}\}$ that represent a style (based on their membership weight $H(k,z)$). Each image will lead to one \snamebig signal, which we average together to obtain the \emph{\snamebig style signal}. This process is repeated for all the dataset partitions presented in FF. To deal with the second challenge, we adopt the best performing statistical forecasting techniques from Fashion Forward and feed them the style \snamebig signal described above. For more details on the forecasting techniques, we refer the reader to ~\cite{FPAP2,TSA}:
\begin{enumerate}
    \item \textbf{Naive methods.} These methods infer by utilizing general information from the training data. \emph{Mean} forecasts the future as the mean of past observations, while \emph{Last} as the last observed value. \emph{Drift} is the same as \emph{Last}, but the forecasts change over time based on the global trend of the series;
    \item \textbf{Auto Regressive and Moving Average methods.} These methods forecast using a linear combination of \emph{some} past observations in a regression framework. The most famous and representative method of this class is the \emph{AutoRegressive Integrated Moving Average}(ARIMA) model.
    \item \textbf{Simple Exponential Smoothing.} Stands for simple exponential smoothing, is a weighted average of previous observations where the weights decrease exponentially as we go further in the past.
\end{enumerate}
Following the protocol of~\cite{al2017fashion}, all models are trained on all but the last timestep, which is used for testing. We utilise the mean absolute percentage error (MAPE) and the mean absolute error (MAE) to evaluate the forecasting accuracy on the last timestep of the signal stemming from FF. To provide an additional comparison, we show additional results using Google Trends as the substitute popularity time series \cite{skenderi2021well}. Note that to obtain fair and comparable results, all the signals are rescaled in the range [0,1] using min-max normalization. The results are shown in Table~\ref{tab:ff_res}, where \textit{Oracle} refers to the original ground-truth style popularity series given as input to the forecasting models. 

 \begin{figure}[t!]
        \centering
        \includegraphics[width=0.48\linewidth]{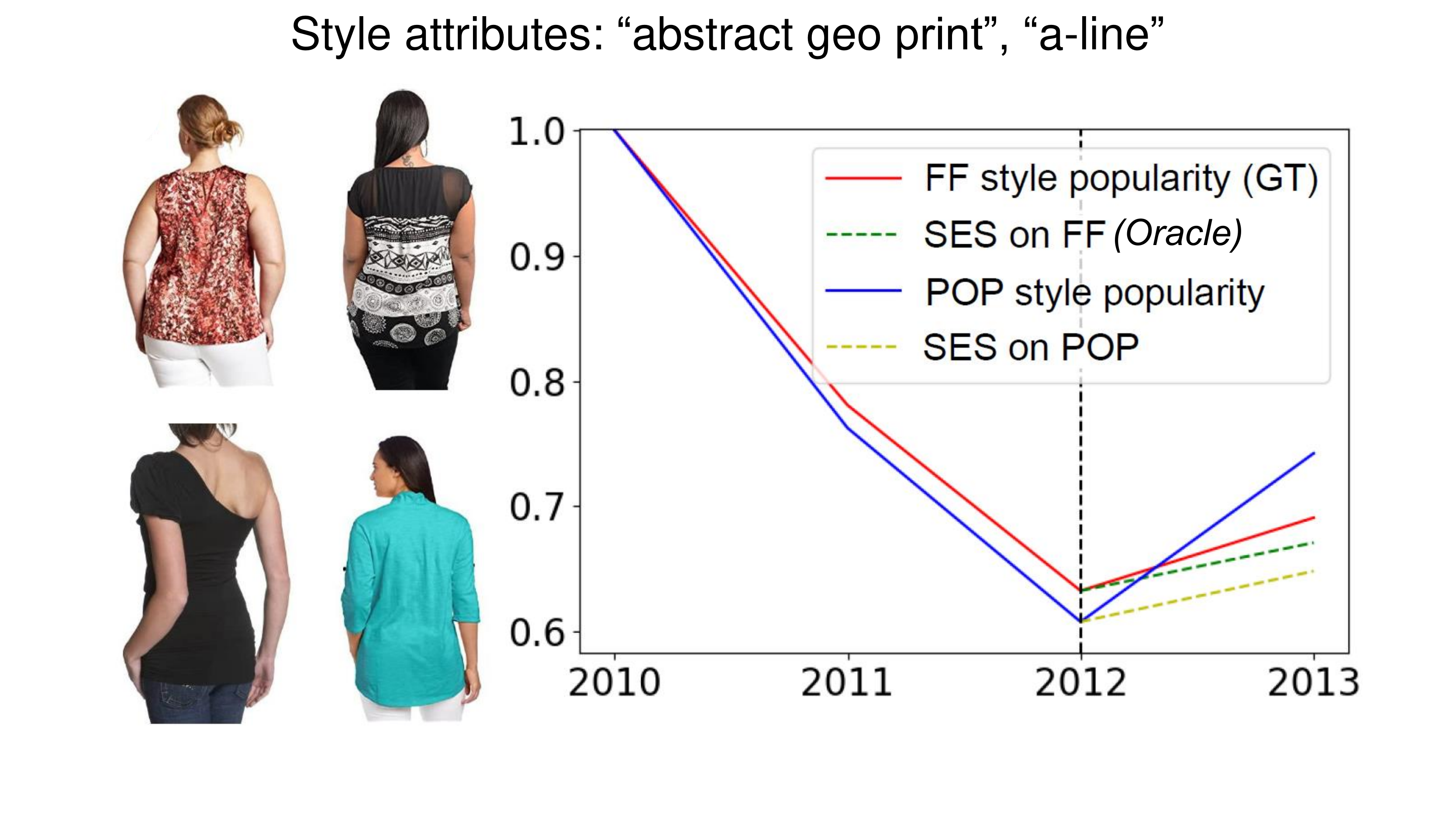}
        \includegraphics[width=0.48\linewidth]{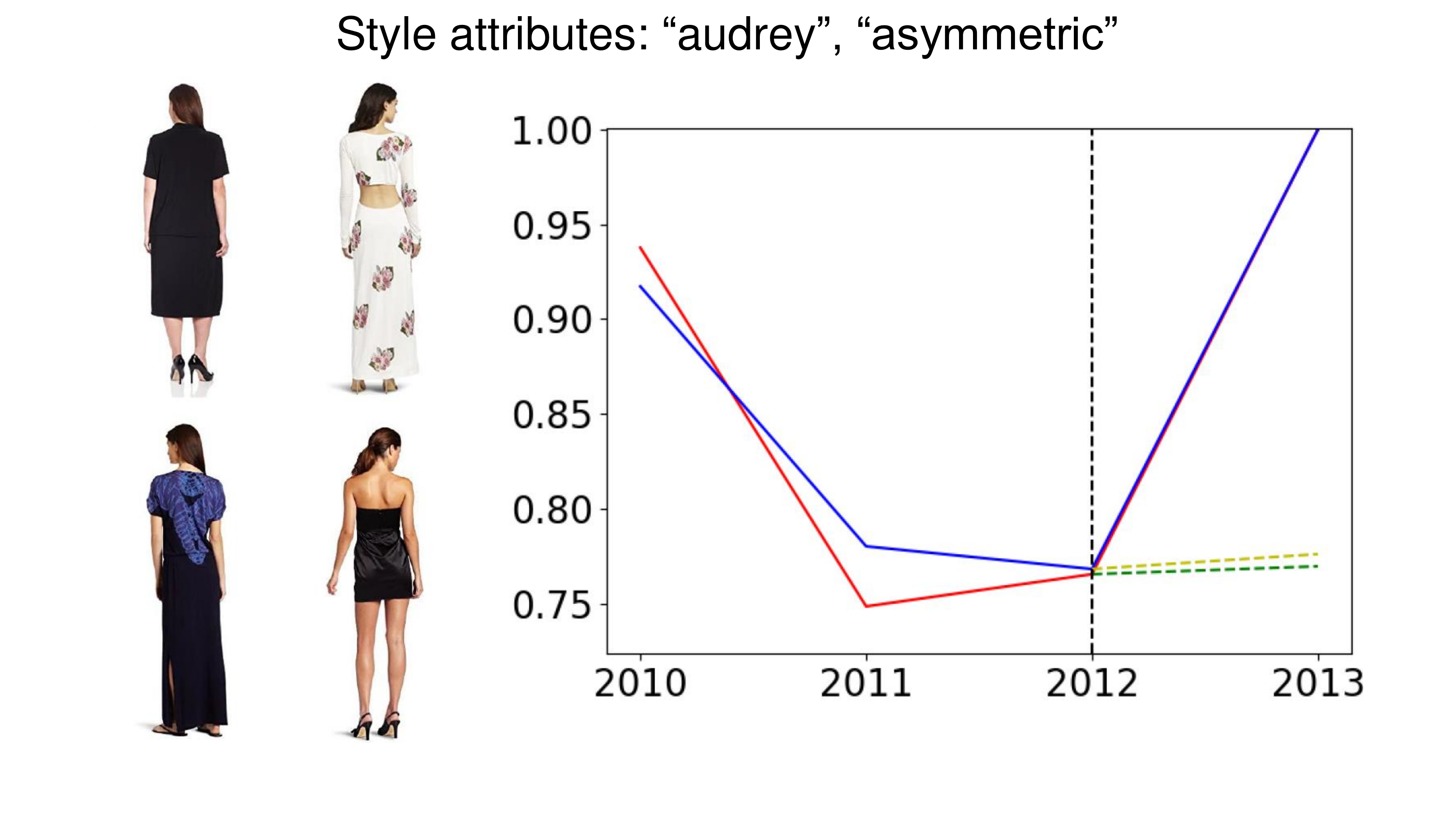}
        \caption{\footnotesize Qualitative results on the forecasting of two different styles from FF, represented by their respective "style-defining" images and top (automatically extracted) attributes.
        In both cases, \snamebig and the ground-truth (GT) style popularity from Fashion Forward are substantially similar. The plot on the right shows a forecasting failure case, which holds for both \snamebig and the GT).}
        \label{fig:qlt_amazon}
        \vspace{-0.5cm}
\end{figure}

\snamebig proves to be a natural substitute to the GT style popularity time series and it allows for optimal forecasts, providing better results than the GT signal itself for the \emph{Dresses} partition. On the other hand, Google Trends are not able to convey such similarities, partially because searching only for the popularity of textual tags might not provide a series that is as predictive as ours.

\section{Conclusion}
\label{sec:conclusion}
Metaphorically, our approach performs a kind of ``time travel'': it sends a fashion probe image in the past, before its launch in the market. It then models the popularity from that past point forward by relying on highly ranked web images, queried by using general textual tags related to the probe. The probe similarity with the past is then shown to be a good exogenous indicator for future performance. This pipeline provides a new, effective and data-centric scheme for NFPPF problems.

\paragraph{\textbf{Acknowledgements.}} This work was supported by the Italian MIUR through the project "Dipartimenti di Eccellenza 2018-2022". 

\section{Appendix}
\label{sec:suppl}
\noindent The supplementary material discusses the following topics, in order:
\begin{enumerate}
    \item Additional ablation study on the signal forming step of \snamebig (Sec. \ref{sec:ablations});
    \item Details on the training data and forecasting horizon on the VISUELLE dataset (Sec. \ref{sec:visuelle});
    \item Complete results on all Fashion Foward (FF) dataset partititions for the new style popularity forecasting task (Sec. \ref{sec:ff});
    \item Qualitative demonstrations of the downloaded images from our temporal cross-modal query expansion, along with interesting insights discovered on the fashionability of different categories and colors (Sec. \ref{sec:qualitative});
    \item Inspired by latest, established guidelines in  CV \& ML research, a discussion on ethical, social and economical implications of our work (Sec. \ref{sec:ethical}).
\end{enumerate}

\subsection{Additional POP ablations}\label{sec:ablations}
This section provides an additional ablation related to the ones shown in Sec. 4.1 and Tab. 3 in the main paper, concerning the \textbf{signal forming step} of the \snamebig signal. This is done by varying the embeddings of the cleaned images by CL that are used as input to Equation 3 (Sec. 3.5 in the main paper), based on the following strategies: 
\begin{itemize}[noitemsep, leftmargin=*]
    \item{\emph{Negative}}: it indicates the average distance of $\mathbf{z}$ with the \emph{pruned unfashionable images} ${}^{(-)}\{\mathbf{x}'_{i}\}_{i=1,\ldots,M''^{(t-k)}}^{(t-k)}$, substituting the positive ones from Equation 3;
    \item{\emph{Positive and Negative}}: here we fed into the forecasting approach two signals, the original POP and the \emph{Negative} one, making POP a multivariate (2D) time series.
\end{itemize}
We discover that the \emph{Negative} approach gives some boost, probably accounting for how much the probe has to be dissimilar to unfashionable items. On the other hand, \emph{Positive and Negative} shows a decrease in comparison to using only the pruned fashionable images, probably because the two signals are complementary (Tab. \ref{tab:alternative_pipelines_supp}). Nevertheless, all the ablated approaches still provide an exogenous time series that helps models perform better when they consider it as additional input.
\begin{table}[h]
\scriptsize
    \centering
    \caption{ Alternative versions of the signal forming step, comparing to the one proposed in Sec. 3.5, represented here as \snamebig, on both the \emph{first order setup} and \emph{release setup} of VISUELLE. Lower is better for both metrics.}
    \setlength\extrarowheight{1pt}
    \resizebox{0.6\linewidth}{!}{
    \begin{tabular}{|l||c|c||c|c|}
        \hline
        &\multicolumn{2}{l||}{\hfil\textbf{\emph{First Order Setup}}}&\multicolumn{2}{l|}{\hfil\textbf{\emph{Release Setup}}} \\
       \textbf{Strategy} & \textbf{WAPE} & \textbf{MAE}& \textbf{WAPE} & \textbf{MAE} \\ \hline
       \emph{Negative}&52.68&28.78&53.90&29.44\\ \hline
        \emph{Positive and Negative}&52.97&28.94&54.35&29.69\\ \hline
        \textbf{POP}&\textbf{52.39}&\textbf{28.62}&\textbf{53.41}&\textbf{29.18}\\ \hline
    \end{tabular}}
    
    \label{tab:alternative_pipelines_supp}
\end{table}

\subsection{Additional details on VISUELLE}\label{sec:visuelle}
\begin{figure}[h!]
    \centering
    \includegraphics[width=0.65\linewidth]{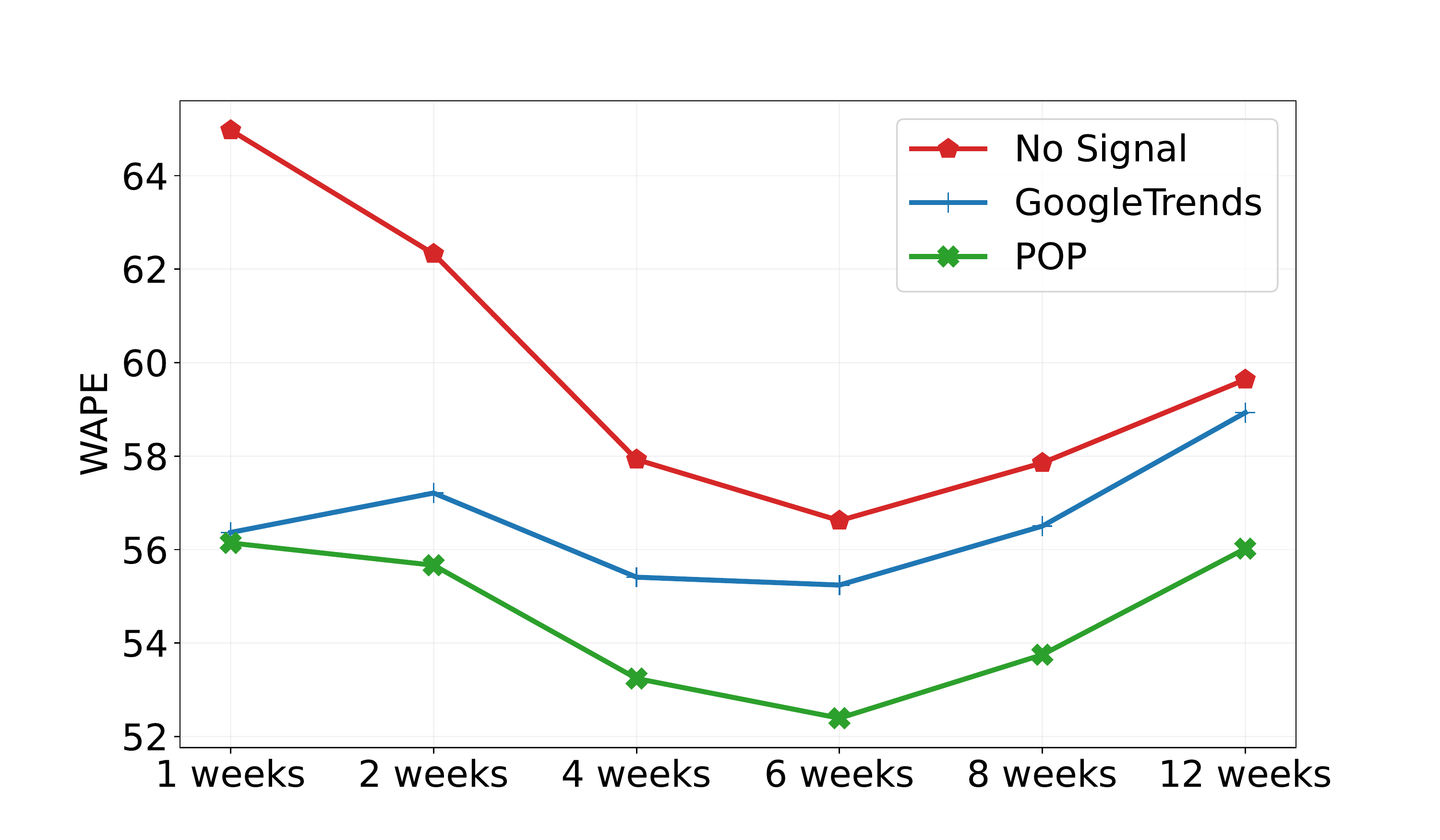}
    \caption{WAPE for different forecasting horizons and exogenous signals, using GTM-Transformer \cite{skenderi2021well} on the VISUELLE dataset. After six weeks there is a long enough history to model tendencies in the sales without considering product discounts or replenishments, unlike longer horizons. This is also reflected in the WAPE values, which keep increasing for forecasting horizons longer than six weeks. POP improves the forecasts for any horizon.}
    \label{fig:result_horizon}
\end{figure}

The VISUELLE dataset from~\cite{skenderi2021well} contains the following, multi-modal information for each product: i) Images, ii) Text tags, iii) Google Trends, iv) Sales curves. 
 \begin{itemize}[noitemsep, leftmargin=*]
     \item{\textbf{Images}}: RGB images with an average size of $577\times 227$; 
     \item{\textbf{Text}}: Three types of text tags (category, dominant color, fabric) that come partially from the technical sheet of the product (category, fabric), while the dominant color has been automatically extracted by the authors of the dataset.
     \item{\textbf{Google Trends}}: Three 52-step long time series for each product, one for each of the tags listed above, extracted by the Google Trend platform in robust way~\cite{medeiros2021proper} for the 52 weeks prior the sale start date;
     \item{\textbf{Sales curve}}: 12-step long weekly time series, reporting the sales of a particular item.
 \end{itemize}
 The testing product set is composed of the 497 most novel products of the two most recent seasons (Spring-Summer and Autumn-Winter 2019). The rest of the dataset (5080 products) is used for training. We \emph{utilise all of this data} for the experiments in Sec 4.1 of the main paper.
 \textbf{Additional forecasting results} are displayed in Fig. \ref{fig:result_horizon} (in terms of WAPE), exhibiting the behaviour of the best performing model for all possible forecasting horizons.
 
 \subsection{Complete results on Fashion Forward}\label{sec:ff}
We show the full new style popularity forecasting results on all dataset partitions of Fashion Forward in Tab. \ref{tab:ff_res_supp}, providing an extended version of Tab. 4 from the main text. The results show how \snamebig can be used to estimate the intra dataset popularity trends of products from Fashion Forward with relatively high accuracy for all partitions. We would like to emphasize again that we use \snamebig as input to forecasting models and compare the predictions with the ground-truth testing set that comes from the FF time series.
\begin{table}[h!]
    \centering
    \caption{Results across all the Fashion Forward~\cite{al2017fashion} datasets.}

    \setlength\extrarowheight{3.5pt}
    \resizebox{\linewidth}{!}{
    \begin{tabular}{|l|c c|c c|c c|c c|c c|c c|}
    \hline
    \multicolumn{13}{|l|}{\hfil\textbf{Global Average}}\\
    \hline
    \textbf{Signals}&\multicolumn{2}{l|}{\hfil\textbf{Mean}}&\multicolumn{2}{l|}{\hfil\textbf{Last}}&\multicolumn{2}{l|}{\hfil\textbf{Drift}}&\multicolumn{2}{l|}{\hfil\textbf{AR}}&\multicolumn{2}{l|}{\hfil\textbf{ARIMA}}&\multicolumn{2}{l|}{\hfil\textbf{SES}}\\
    &MAE&MAPE&MAE&MAPE&MAE&MAPE&MAE&MAPE&MAE&MAPE&MAE&MAPE\\ \hline
    \emph{Oracle}&\emph{0.136}&\emph{0.170}&\emph{0.093}&\emph{0.114}&\emph{0.174}&\emph{0.222}&\emph{0.271}&\emph{0.403}&\emph{0.136}&\emph{0.167}&\emph{0.094}&\emph{0.116}\\\hline
    GoogleTrends&0.846&1.000&0.846&1.000&0.846&1.000&0.846&1.000&0.846&1.000&0.846&1.000\\ \hline
    \textbf{POP}&\textbf{0.152}&\textbf{0.192}&\textbf{0.116}&\textbf{0.144}&\textbf{0.182}&\textbf{0.229}&\textbf{0.281}&\textbf{0.418}&\textbf{0.235}&\textbf{0.293}&\textbf{0.125}&\textbf{0.156}\\\hline\hline
    \multicolumn{13}{|l|}{\hfil\textbf{Dresses}}\\
    \hline
    \textbf{Signals}&\multicolumn{2}{l|}{\hfil\textbf{Mean}}&\multicolumn{2}{l|}{\hfil\textbf{Last}}&\multicolumn{2}{l|}{\hfil\textbf{Drift}}&\multicolumn{2}{l|}{\hfil\textbf{AR}}&\multicolumn{2}{l|}{\hfil\textbf{ARIMA}}&\multicolumn{2}{l|}{\hfil\textbf{SES}}\\
    &MAE&MAPE&MAE&MAPE&MAE&MAPE&MAE&MAPE&MAE&MAPE&MAE&MAPE\\ \hline
    \emph{Oracle}&\emph{0.155}&\emph{0.197}&\emph{0.130}&\emph{0.158}&\emph{0.203}&\emph{0.263}&\emph{0.307}&\emph{0.409}&\emph{0.173}&\emph{0.209}&\emph{0.129}&\emph{0.157}\\\hline
    GoogleTrends&0.849&1.000&0.849&1.000&0.849&1.000&0.849&1.000&0.849&1.000&0.849&1.000\\\hline
    \textbf{POP}&\textbf{0.119}&\textbf{0.157}&\textbf{0.108}&\textbf{0.127}&\textbf{0.173}&\textbf{0.216}&\textbf{0.229}&\textbf{0.334}&\textbf{0.162}&\textbf{0.193}&\textbf{0.109}&\textbf{0.130}\\\hline\hline
    \multicolumn{13}{|l|}{\hfil\textbf{Shirts}}\\
    \hline
    \textbf{Signals}&\multicolumn{2}{l|}{\hfil\textbf{Mean}}&\multicolumn{2}{l|}{\hfil\textbf{Last}}&\multicolumn{2}{l|}{\hfil\textbf{Drift}}&\multicolumn{2}{l|}{\hfil\textbf{AR}}&\multicolumn{2}{l|}{\hfil\textbf{ARIMA}}&\multicolumn{2}{l|}{\hfil\textbf{SES}}\\
    &MAE&MAPE&MAE&MAPE&MAE&MAPE&MAE&MAPE&MAE&MAPE&MAE&MAPE\\ \hline
    \emph{Oracle}&\emph{0.122}&\emph{0.149}&\emph{0.075}&\emph{0.097}&\emph{0.148}&\emph{0.190}&\emph{0.301}&\emph{0.371}&\emph{0.126}&\emph{0.159}&\emph{0.080}&\emph{0.103}\\\hline
    GoogleTrends&0.840&1.000&0.840&1.000&0.840&1.000&0.840&1.000&0.840&1.000&0.840&1.000\\\hline
    \textbf{POP}&\textbf{0.144}&\textbf{0.175}&\textbf{0.109}&\textbf{0.152}&\textbf{0.166}&\textbf{0.215}&\textbf{0.274}&\textbf{0.336}&\textbf{0.139}&\textbf{0.189}&\textbf{0.111}&\textbf{0.151}\\ \hline\hline
    \multicolumn{13}{|l|}{\hfil\textbf{Tops\&Tees}}\\
    \hline
    \textbf{Signals}&\multicolumn{2}{l|}{\hfil\textbf{Mean}}&\multicolumn{2}{l|}{\hfil\textbf{Last}}&\multicolumn{2}{l|}{\hfil\textbf{Drift}}&\multicolumn{2}{l|}{\hfil\textbf{AR}}&\multicolumn{2}{l|}{\hfil\textbf{ARIMA}}&\multicolumn{2}{l|}{\hfil\textbf{SES}}\\
    &MAE&MAPE&MAE&MAPE&MAE&MAPE&MAE&MAPE&MAE&MAPE&MAE&MAPE\\ \hline
    \emph{Oracle}&\emph{0.132}&\emph{0.165}&\emph{0.074}&\emph{0.087}&\emph{0.172}&\emph{0.212}&\emph{0.206}&\emph{0.429}&\emph{0.108}&\emph{0.133}&\emph{0.073}&\emph{0.087}\\\hline
    GoogleTrends&0.848&1.000&0.848&1.000&0.848&1.000&0.848&1.000&0.848&1.000&0.848&1.000\\\hline
    \textbf{POP}&\textbf{0.193}&\textbf{0.245}&\textbf{0.131}&\textbf{0.153}&\textbf{0.206}&\textbf{0.257}&\textbf{0.341}&\textbf{0.585}&\textbf{0.405}&\textbf{0.497}&\textbf{0.156}&\textbf{0.186}\\\hline
    \end{tabular}}  
    \label{tab:ff_res_supp}
\end{table}

\begin{figure}[t!]
    \begin{center}
            \includegraphics[width=\linewidth]{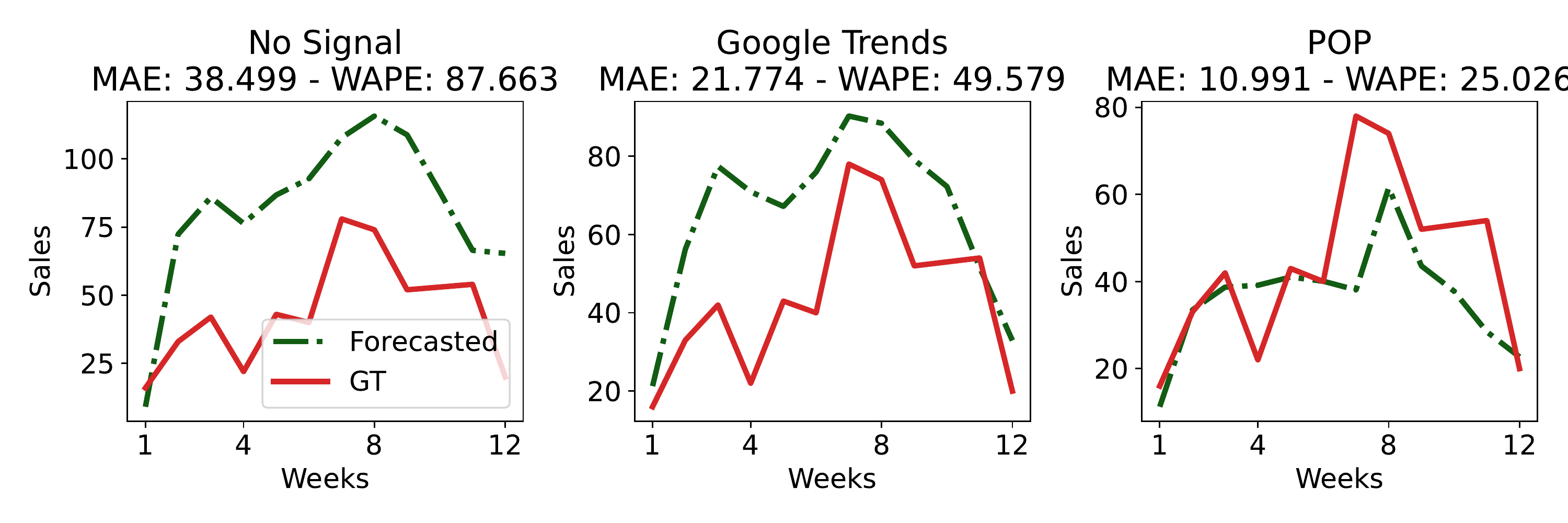}
            \includegraphics[width=\linewidth]{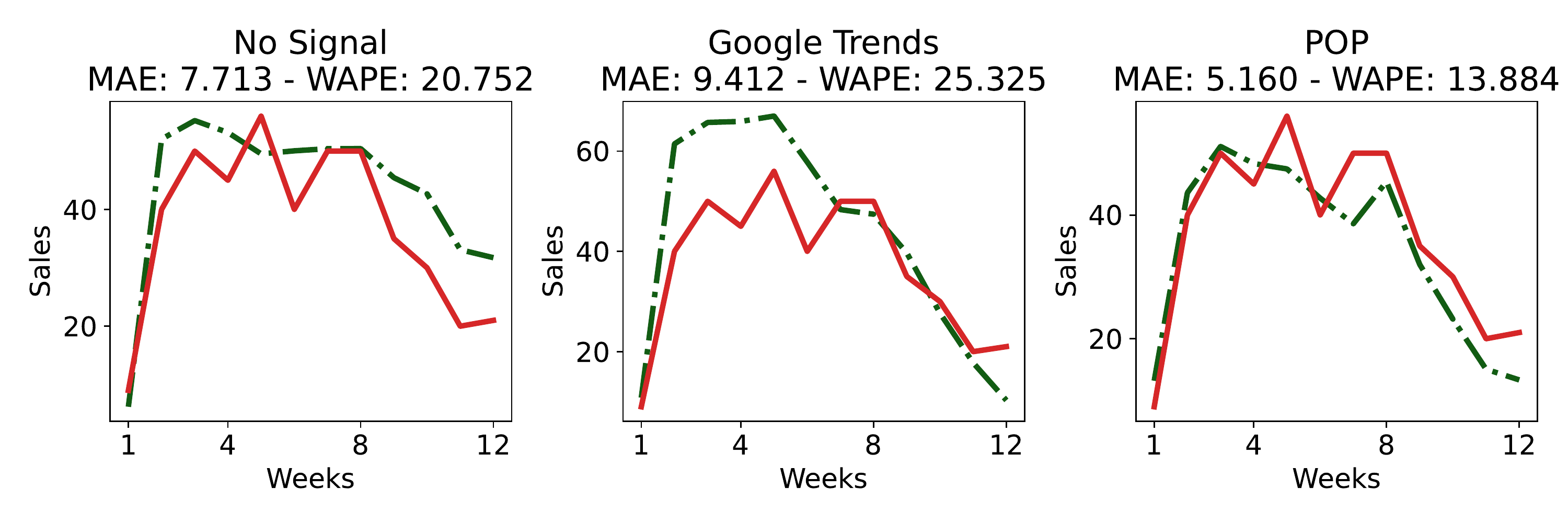}
            \includegraphics[width=\columnwidth]{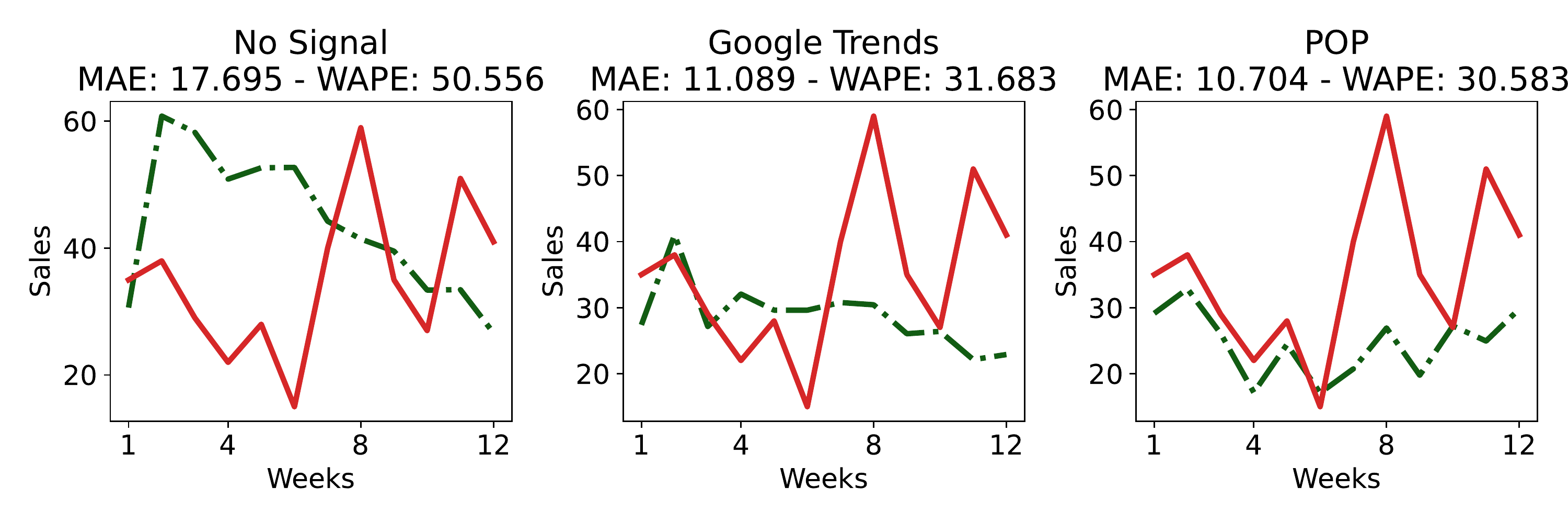}
    \end{center}
    \caption{\label{fig:qlt_visuelle_supp} Qualitative results on VISUELLE, considering all the 12 time-steps. In all the cases, using POP gives better results than not using it or using other, similar exagenous series.}
\end{figure}

\subsection{Qualitative Results}\label{sec:qualitative}

In this section we report a qualitative analysis of the POP signal, which in the main paper was limited to Fig.3 due to space limitations. These results give additional insight on the significance of our time-dependent, data-centric approach.

\subsection{Under the hood of the POP signal}
\begin{figure}[t!]
    \begin{center}
        \includegraphics[width=\linewidth]{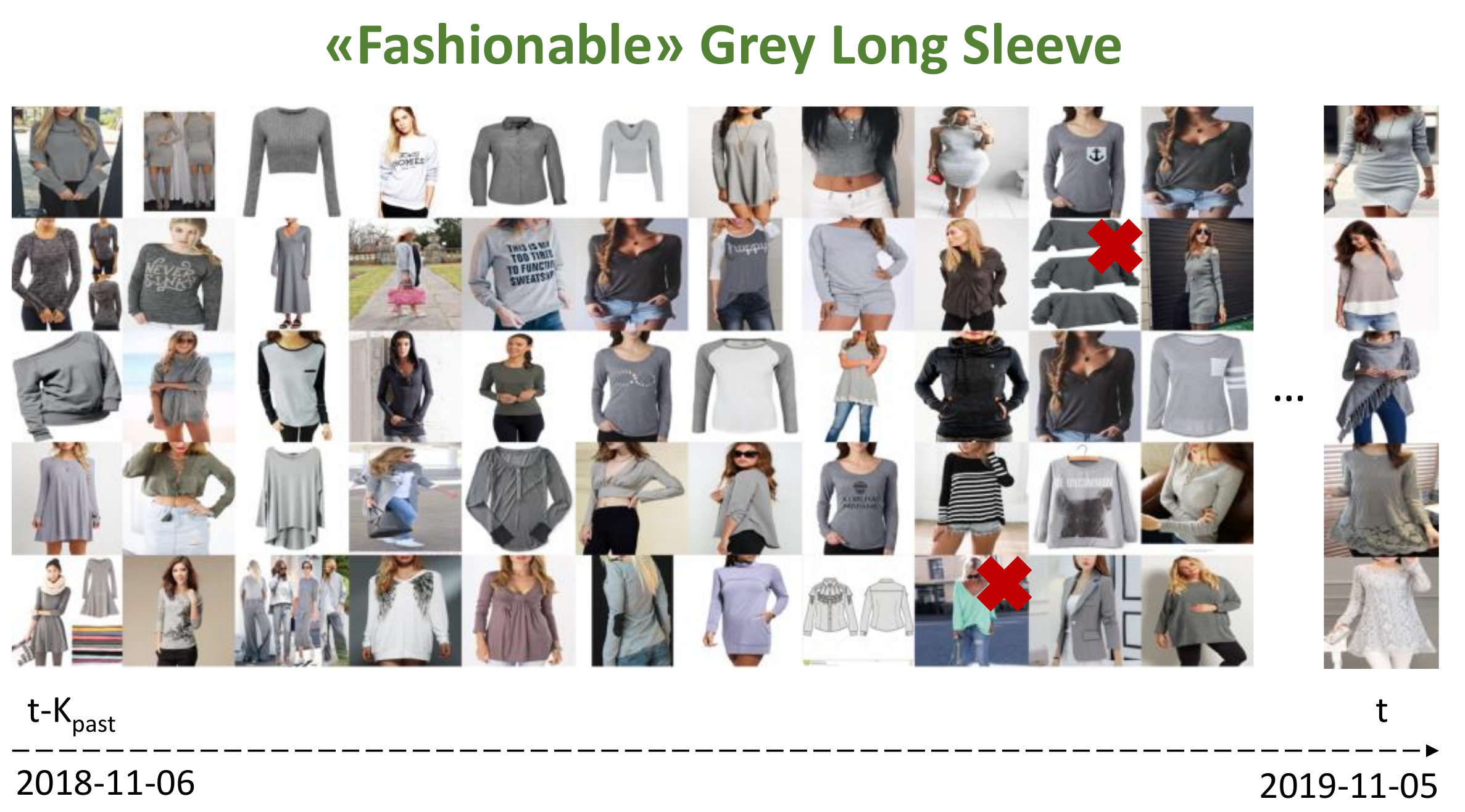}
        \includegraphics[width=\linewidth]{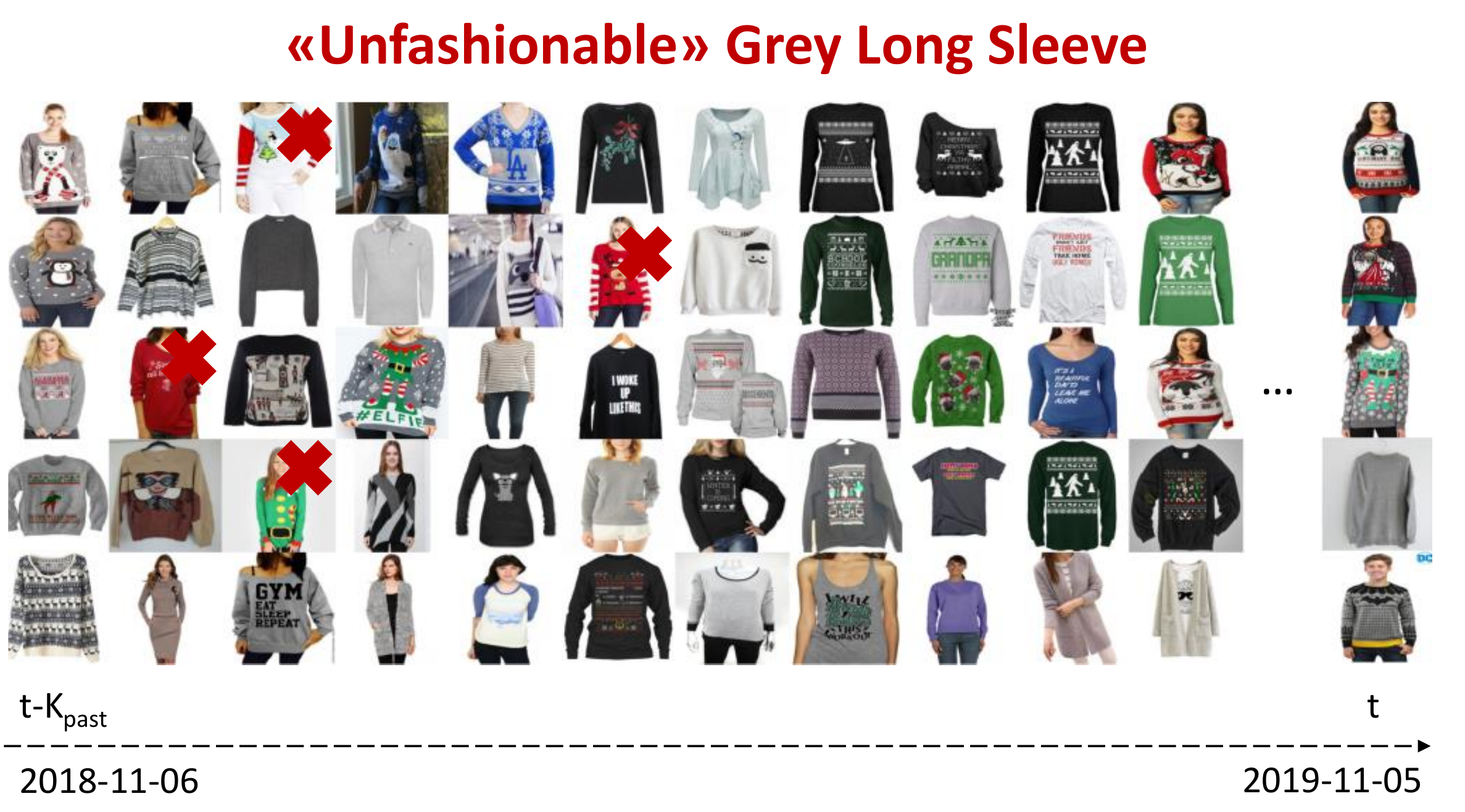}
    \end{center}
    \caption{Examples of images downloaded for the query `Grey Long Sleeves'' (after pruning by confident learning). One may note that mismatching images are very few, intended as those images which are not containing any "Grey Long Sleeves". An example would be the green sleeve + blue jeans in the bottom row. It is worth noting how most of the fashionable items have no printed logos, texture or tight sleeves. On the contrary,``Unfashionable Grey Long Sleeves'' have big logo on them, with a winter theme, and many colors accompanying a gray background. In some cases, the gray color actually covers a small portion of the clothing item. Pruned images are marked with a red cross.}
    \label{fig:grey}
\end{figure}

\begin{figure}[]
    \begin{center}
        \includegraphics[width=\linewidth]{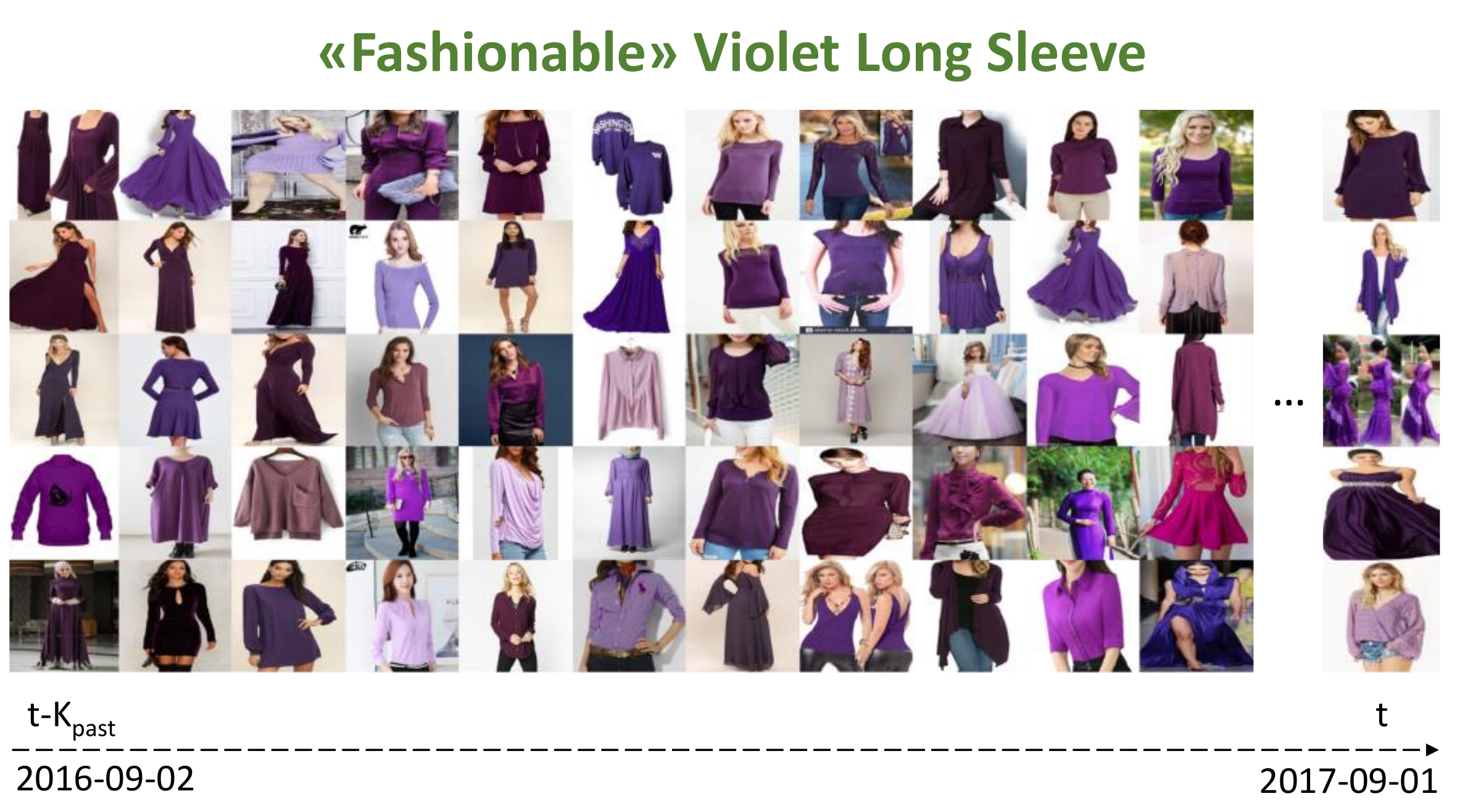}
        \includegraphics[width=\linewidth]{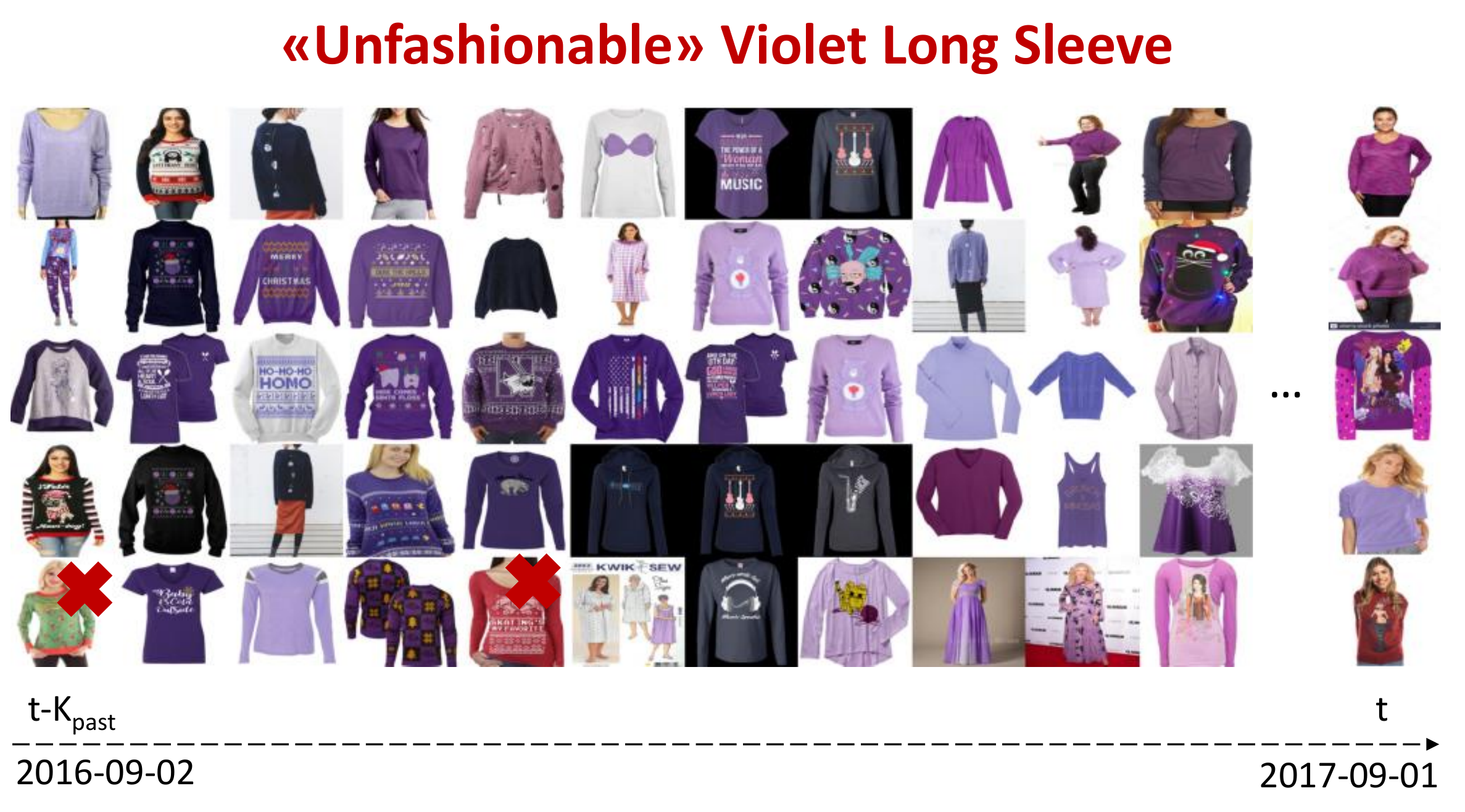}
    \end{center}
    \caption{Examples of images downloaded for the query `Violet Long Sleeve'' (after pruning by confident learning). The ``Fashionable Violet Long Sleeve'' items seem to have a darker tone in most cases. Very long sleeves fade into dresses, indicating the length of the garment as an important aspect for making it fashionable. Curiously, ``Unfashionable Violet Long Sleeve'' contain brighter colors, short garments (like pyjamas) with writings or printed images. Pruned images are marked with a red cross.}
    \label{fig:viol}
\end{figure}

In Fig.~\ref{fig:grey} and Fig.~\ref{fig:viol} we report two examples of the (automatically) downloaded images used for the formation of the \snamebig signal. In both figures, the probe images from which we extract the textual attributes to index the search are depicted. The analysis for each figure is reported in the corresponding caption. We also report in the figures some pruned images by the confident learning step, marked by a red cross. 


\subsection{The importance of a time-dependent query}
\begin{figure}[]
    \begin{center}
        \includegraphics[width=\linewidth]{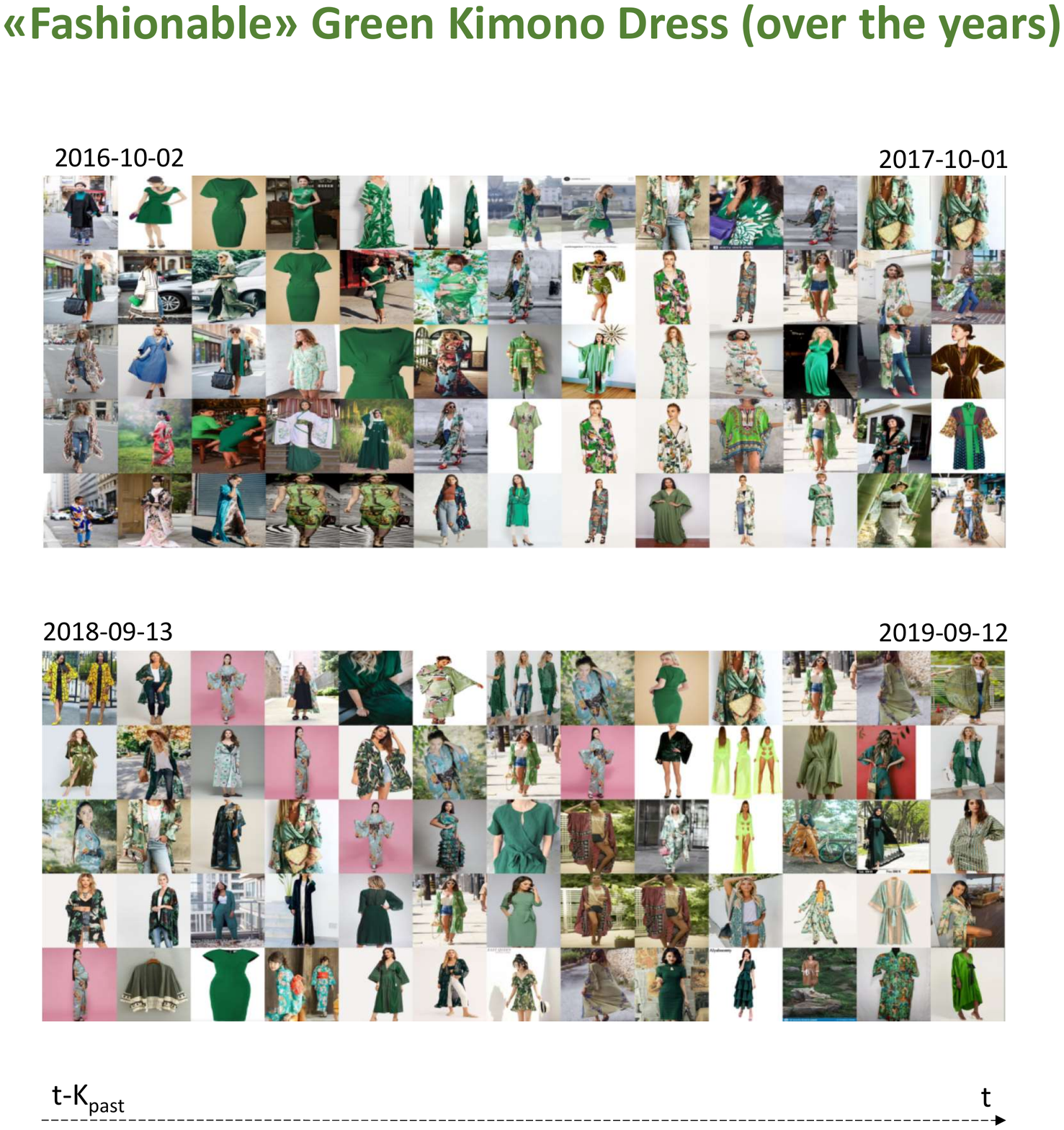}
    \end{center}
    \caption{Examples of Fashionable downloaded images for particular time-depended queries. In this particular case, for the query "green kimono dress", it can be seen how the notion of fashionability can have significant variations over time. Notably, green kimonos in 2017, as seen in the latter half of the first figure, tend to be heavily associated with white patterns and the color white in general. In 2019, this trend appears to be dying out, with the kimonos being of different shades of green or even dark green.}
    \label{fig:kimono} 
\end{figure}

In the main paper, we report in Tab. 3 various ablation studies on POP. Here we focus on two aspects in particular: the ``Time Dependent Query Expansion'', and the ``misaligned Past''. The obtained results suggest that exploiting (un)fashionable images not related to the date of delivery on the market gives worse results in terms of forecasting. Fig.~\ref{fig:kimono} qualitatively demonstrates why this is the case. As it is visible, what made a garment of a particular type and color fashionable in 2017 (Fig.~\ref{fig:kimono}, top) does not correspond to the same visual elements that can be found in 2019 (Fig.~\ref{fig:kimono}, bottom). More specifically, throughout the spring/summer season of 2017, the green kimonos tend to be heavily associated with white patterns and the color white in general. In 2019, the kimonos are almost all in different shades of green or even dark green.

\subsection{Ethical Concerns and Societal Impact} \label{sec:ethical} 
\textbf{Ethical implications} could, in principle, arise from the web image search: observed images can, for example, contain copyrighted images. Nevertheless, just as a normal user would use Google Images to gather an opinion of what could be trending in fashion, so do we, albeit automatically. In particular, we do not need to personally look at the web images (apart for the ones shown in the paper and here for explanatory purposes), since \snamebig is a just a numerical time series. \textbf{As for the societal impact}, our approach can be highly beneficial for fast fashion, which is the third most polluting industry in the world. Having a precise estimation of sales or popularity can improve the situation by solving supply chain issues and our pipeline can play a leading part. Ameliorated forecasts with \snamebig also have a big impact at the economic level, in terms of profit. The best forecasting model on the VISUELLE dataset (which uses our generated time series, as seen in Tables 1 and 2 in the main paper), allows to spare 21\% w.r.t. ordinary guidelines for new fast-fashion products, reducing a loss of \$4.390.400 US dollars to \$3.491.600 US dollars, assuming a general price of 28\$ per piece for all products (independently on the category).

\clearpage
%
%
\bibliographystyle{splncs04}
\bibliography{egbib}

\begin{thebibliography}{10}
\providecommand{\url}[1]{\texttt{#1}}
\providecommand{\urlprefix}{URL }
\providecommand{\doi}[1]{https://doi.org/#1}

\bibitem{al2017fashion}
Al-Halah, Z., Stiefelhagen, R., Grauman, K.: Fashion forward: Forecasting
  visual style in fashion. In: ICCV (2017)

\bibitem{anik2021data}
Anik, A.I., Bunt, A.: Data-centric explanations: Explaining training data of
  machine learning systems to promote transparency. In: Proceedings of the 2021
  CHI Conference on Human Factors in Computing Systems (2021)

\bibitem{arvan2019integrating}
Arvan, M., Fahimnia, B., Reisi, M., Siemsen, E.: Integrating human judgement
  into quantitative forecasting methods: A review. Omega  \textbf{86} (2019)

\bibitem{TSA}
Box, G., Jenkins, G., Reinsel, G., Ljung, G.: Time Series Analysis: Forecasting
  and Control. John Wiley \& Sons (2015)

\bibitem{chen2004marriage}
Chen, L., Ng, R.: On the marriage of lp-norms and edit distance. In:
  Proceedings of the Thirtieth international conference on Very large data
  bases-Volume 30 (2004)

\bibitem{chen_webly_2015}
Chen, X., Gupta, A.: Webly supervised learning of convolutional networks. In:
  2015 IEEE International Conference on Computer Vision (ICCV). pp. 1431--1439.
  IEEE Computer Society, Los Alamitos, CA, USA (dec 2015).
  \doi{10.1109/ICCV.2015.168},
  \url{https://doi.ieeecomputersociety.org/10.1109/ICCV.2015.168}

\bibitem{cheng2021fashion}
Cheng, W.H., Song, S., Chen, C.Y., Hidayati, S.C., Liu, J.: Fashion meets
  computer vision: A survey. ACM Computing Surveys (CSUR)  \textbf{54}(4)
  (2021)

\bibitem{imagenet}
Deng, J., Dong, W., Socher, R., Li, L.J., Li, K., Fei-Fei, L.: Imagenet: A
  large-scale hierarchical image database. In: 2009 IEEE Conference on Computer
  Vision and Pattern Recognition (2009). \doi{10.1109/CVPR.2009.5206848}

\bibitem{ekambaram_attention_2020}
Ekambaram, V., Manglik, K., Mukherjee, S., Sajja, S.S.K., Dwivedi, S., Raykar,
  V.: Attention based {Multi}-{Modal} {New} {Product} {Sales} {Time}-series
  {Forecasting}. In: Proceedings of the 26th {ACM} {SIGKDD} {International}
  {Conference} on {Knowledge} {Discovery} \& {Data} {Mining}. ACM, Virtual
  Event CA USA (Aug 2020). \doi{10.1145/3394486.3403362},
  \url{https://dl.acm.org/doi/10.1145/3394486.3403362}

\bibitem{Fergus_googleimages_2005}
Fergus, R., Fei-Fei, L., Perona, P., Zisserman, A.: Learning object categories
  from google's image search. In: Tenth IEEE International Conference on
  Computer Vision (ICCV'05) Volume 1. vol.~2, pp. 1816--1823 Vol. 2 (2005).
  \doi{10.1109/ICCV.2005.142}

\bibitem{fildes2019retail}
Fildes, R., Ma, S., Kolassa, S.: Retail forecasting: Research and practice.
  International Journal of Forecasting  (2019)

\bibitem{garcia2021fashion}
Garcia, C.C.: Fashion forecasting: an overview from material culture to
  industry. Journal of Fashion Marketing and Management: An International
  Journal  (2021)

\bibitem{he2015deep}
He, K., Zhang, X., Ren, S., Sun, J.: Deep residual learning for image
  recognition (2015)

\bibitem{FPAP2}
Hyndman, R., Athanasopoulos, G.: Forecasting: Principles and Practice. OTexts,
  Australia, 2nd edn. (2018)

\bibitem{ilic2021explainable}
Ilic, I., G{\"o}rg{\"u}l{\"u}, B., Cevik, M., Baydo{\u{g}}an, M.G.: Explainable
  boosted linear regression for time series forecasting. Pattern Recognition
  (2021)

\bibitem{jeon2020fashionq}
Jeon, Y., Jin, S., Kim, B., Han, K.: Fashionq: An interactive tool for
  analyzing fashion style trend with quantitative criteria. In: Extended
  Abstracts of the 2020 CHI Conference on Human Factors in Computing Systems
  (2020)

\bibitem{google_visualrank}
Jing, Y., Baluja, S.: Visualrank: Applying pagerank to large-scale image
  search. IEEE Transactions on Pattern Analysis and Machine Intelligence
  \textbf{30},  1877--1890 (2008)

\bibitem{Li_webly_2021}
Li, J., Song, Y., Zhu, J., Cheng, L., Su, Y., Ye, L., Yuan, P., Han, S.:
  Learning from large-scale noisy web data with ubiquitous reweighting for
  image classification. IEEE Transactions on Pattern Analysis and Machine
  Intelligence  \textbf{43}(5),  1808--1814 (2021).
  \doi{10.1109/TPAMI.2019.2961910}

\bibitem{liuLQWTcvpr16DeepFashion}
Liu, Z., Luo, P., Qiu, S., Wang, X., Tang, X.: Deepfashion: Powering robust
  clothes recognition and retrieval with rich annotations. In: Proceedings of
  IEEE Conference on Computer Vision and Pattern Recognition (CVPR) (June 2016)

\bibitem{lo_dressing_2019}
Lo, L., Liu, C., Lin, R., Wu, B., Shuai, H., Cheng, W.: Dressing for
  {Attention}: {Outfit} {Based} {Fashion} {Popularity} {Prediction}. In: 2019
  {IEEE} {International} {Conference} on {Image} {Processing} ({ICIP}) (Sep
  2019). \doi{10.1109/ICIP.2019.8803461}, iSSN: 2381-8549

\bibitem{loshchilov2018decoupled}
Loshchilov, I., Hutter, F.: Decoupled weight decay regularization. In:
  International Conference on Learning Representations (2018)

\bibitem{ma_knowledge_2020}
Ma, Y., Ding, Y., Yang, X., Liao, L., Wong, W.K., Chua, T.S.: Knowledge
  {Enhanced} {Neural} {Fashion} {Trend} {Forecasting}. In: Proceedings of the
  2020 {International} {Conference} on {Multimedia} {Retrieval}. ACM, Dublin
  Ireland (Jun 2020). \doi{10.1145/3372278.3390677},
  \url{https://dl.acm.org/doi/10.1145/3372278.3390677}

\bibitem{ma2020kern}
Ma, Y., Ding, Y., Yang, X., Liao, L., Wong, W.K., Chua, T.S.: Knowledge
  enhanced neural fashion trend forecasting. In: Proceedings of the 2020
  International Conference on Multimedia Retrieval. ICMR '20, Association for
  Computing Machinery, New York, NY, USA (2020). \doi{10.1145/3372278.3390677},
  \url{https://doi.org/10.1145/3372278.3390677}

\bibitem{2015_amazonreviews}
McAuley, J., Targett, C., Shi, Q., van~den Hengel, A.: Image-based
  recommendations on styles and substitutes. In: Proceedings of the 38th
  International ACM SIGIR Conference on Research and Development in Information
  Retrieval. p. 43–52. SIGIR '15, Association for Computing Machinery, New
  York, NY, USA (2015). \doi{10.1145/2766462.2767755},
  \url{https://doi.org/10.1145/2766462.2767755}

\bibitem{medeiros2021proper}
Medeiros, M.C., Pires, H.F.: The proper use of google trends in forecasting
  models (2021)

\bibitem{motamedi2021data}
Motamedi, M., Sakharnykh, N., Kaldewey, T.: A data-centric approach for
  training deep neural networks with less data. arXiv preprint arXiv:2110.03613
   (2021)

\bibitem{NG:2021}
Ng, A.: A chat with andrew on mlops: From model-centric to data-centric ai.
  \url{https://www.youtube.com/watch?v=06-AZXmwHjo} (May 2021)

\bibitem{northcutt2021confident}
Northcutt, C., Jiang, L., Chuang, I.: Confident learning: Estimating
  uncertainty in dataset labels. Journal of Artificial Intelligence Research
  \textbf{70} (2021)

\bibitem{northcutt2021pervasive}
Northcutt, C.G., ChipBrain, M., Athalye, A., Mueller, J.: Pervasive label
  errors in test sets destabilize machine learning benchmarks. stat
  \textbf{1050} (2021)

\bibitem{ren2017comparative}
Ren, S., Chan, H.L., Ram, P.: A comparative study on fashion demand forecasting
  models with multiple sources of uncertainty. Annals of Operations Research
  \textbf{257}(1) (2017)

\bibitem{ren2020demand}
Ren, S., Chan, H.L., Siqin, T.: Demand forecasting in retail operations for
  fashionable products: methods, practices, and real case study. Annals of
  Operations Research  \textbf{291}(1) (2020)

\bibitem{shazeer2018adafactor}
Shazeer, N., Stern, M.: Adafactor: Adaptive learning rates with sublinear
  memory cost. In: Dy, J., Krause, A. (eds.) Proceedings of the 35th
  International Conference on Machine Learning. Proceedings of Machine Learning
  Research, vol.~80. PMLR (10--15 Jul 2018),
  \url{https://proceedings.mlr.press/v80/shazeer18a.html}

\bibitem{silva2019googling}
Silva, E.S., Hassani, H., Madsen, D.{\O}., Gee, L.: Googling fashion:
  forecasting fashion consumer behaviour using google trends. Social Sciences
  \textbf{8}(4) (2019)

\bibitem{singh_fashion_2019}
Singh, P.K., Gupta, Y., Jha, N., Rajan, A.: Fashion {Retail}: {Forecasting}
  {Demand} for {New} {Items}. In: 26th ACM SIGKDD Conference on Knowledge
  Discovery and Data Mining (Jun 2019), \url{http://arxiv.org/abs/1907.01960}

\bibitem{skenderi2021well}
Skenderi, G., Joppi, C., Denitto, M., Cristani, M.: Well googled is half done:
  Multimodal forecasting of new fashion product sales with image-based google
  trends. arXiv preprint arXiv:2109.09824  (2021)

\bibitem{song2019selfie}
Song, H., Kim, M., Lee, J.G.: Selfie: Refurbishing unclean samples for robust
  deep learning. In: International Conference on Machine Learning. PMLR (2019)

\bibitem{sorger2017fundamentals}
Sorger, R., Udale, J.: The fundamentals of fashion design. Bloomsbury
  Publishing (2017)

\bibitem{vaswani2017attention}
Vaswani, A., Shazeer, N., Parmar, N., Uszkoreit, J., Jones, L., Gomez, A.N.,
  Kaiser, L., Polosukhin, I.: Attention is all you need (2017)

\bibitem{wang2019symmetric}
Wang, Y., Ma, X., Chen, Z., Luo, Y., Yi, J., Bailey, J.: Symmetric cross
  entropy for robust learning with noisy labels. In: Proceedings of the
  IEEE/CVF International Conference on Computer Vision (2019)

\end{thebibliography}
\end{document}